\documentclass{article} 
\PassOptionsToPackage{table}{xcolor}
\usepackage{iclr2027_conference,times}

\usepackage{url}
\usepackage{amsthm}
\usepackage{amsmath,amsfonts,bm}
\usepackage{mathtools}
\usepackage{todonotes}
\usepackage{booktabs}
\usepackage[table]{xcolor}
\usepackage{multirow}
\usepackage{graphicx}
\usepackage{algorithm}
\usepackage{algorithmicx}
\usepackage{algpseudocode}

\usepackage{hyperref}
\usepackage{cleveref}


\makeatletter
\DeclareRobustCommand\onedot{\futurelet\@let@token\@onedot}
\def\@onedot{\ifx\@let@token.\else.\null\fi\xspace}

\makeatother

\crefname{appendixsec}{Appendix}{Appendices}
\Crefname{appendixsec}{Appendix}{Appendices}

\newcommand{\ignore}[1]{}   

\definecolor{bestpurple}{HTML}{5B2C87}
\newcommand{\best}[1]{\textbf{\textcolor{bestpurple}{\bf{#1}}}}

\newcolumntype{L}[1]{>{\raggedright\let\newline\\\arraybackslash\hspace{0pt}}m{#1}}
\newcolumntype{C}[1]{>{\centering\let\newline\\\arraybackslash\hspace{0pt}}m{#1}}
\newcolumntype{R}[1]{>{\raggedleft\let\newline\\\arraybackslash\hspace{0pt}}m{#1}}

\definecolor{tabfirst}{rgb}{1, 0.7, 0.7} 
\definecolor{tabsecond}{rgb}{1, 0.85, 0.7} 
\definecolor{tabthird}{rgb}{1, 1, 0.7} 

\newtheorem{definition}{Definition}
\newtheorem{proposition}{Proposition}

\def\eqref#1{equation~\ref{#1}}

\def\plaineqref#1{\ref{#1}}

\def\1{\bm{1}}

\DeclareMathAlphabet{\mathsfit}{\encodingdefault}{\sfdefault}{m}{sl}
\SetMathAlphabet{\mathsfit}{bold}{\encodingdefault}{\sfdefault}{bx}{n}

\newcommand{\E}{\mathbb{E}}

\newcommand{\KL}{D_{\mathrm{KL}}}

\DeclareMathOperator*{\argmax}{arg\,max}
\DeclareMathOperator*{\argmin}{arg\,min}

\title{Latent Space Planning with Sequential Monte Carlo}

\newif\ifpreprint
\preprinttrue

\author{Hrishikesh Viswanath \\
Department of Computer Science \\
Purdue University \\
\texttt{hviswan@purdue.edu} \\
\And
Anindita Ghosh \\
Saarland University \\
\AND
Uttaran Bhattacharya \\
Adobe Systems \\
\And
Aniket Bera \\
Department of Computer Science \\
Purdue University
}

\ifpreprint\iclrfinalcopy\fi
\begin{document}

\maketitle
\ifpreprint
\lhead{Preprint. Under review.}
\begin{center}
\vskip -0.15in
\url{https://github.com/HrishikeshVish/smcplan}
\end{center}
\fi

\begin{abstract}
Planning with learned world models amounts to finding action sequences whose predicted rollouts reach a goal. Sampling-based planners often struggle at longer horizons, while gradient-based methods can become trapped in local minima. Meanwhile, model predictive path integral control (MPPI), a complementary approach, alternates between estimating an expectation under a cost-weighted distribution from samples and updating the plan from this estimate. Its samples, however, are drawn afresh at every iteration from a Gaussian proposal and reach the target distribution only through importance weights. Moreover, a target distribution favoring goal-reaching trajectories can be difficult to define directly, motivating a sequence of intermediate distributions. Sequential Monte Carlo (SMC) provides a natural mechanism for evolving such a sequence through importance weighting and resampling. Motivated by this, we formulate latent-space planning as optimization guided by SMC, which moves a population of action sequences while a gradient-based optimizer updates the plan. Inspired by classical planning, we further introduce a PDE-derived cost-to-go that guides the evolution of the sample distribution using the structure of the latent state space. On navigation and manipulation tasks with visual and proprioceptive world models, our formulation outperforms gradient-based planners by $29$ and $50$, sampling-based planners by $35$ and MPPI by $7$ percentage points in average success rate, while succeeding in fewer iterations than gradient-based planners and within 4\% of MPPI.
\end{abstract}

\begin{figure}[!h]
    \centering
    \includegraphics[width=\linewidth]{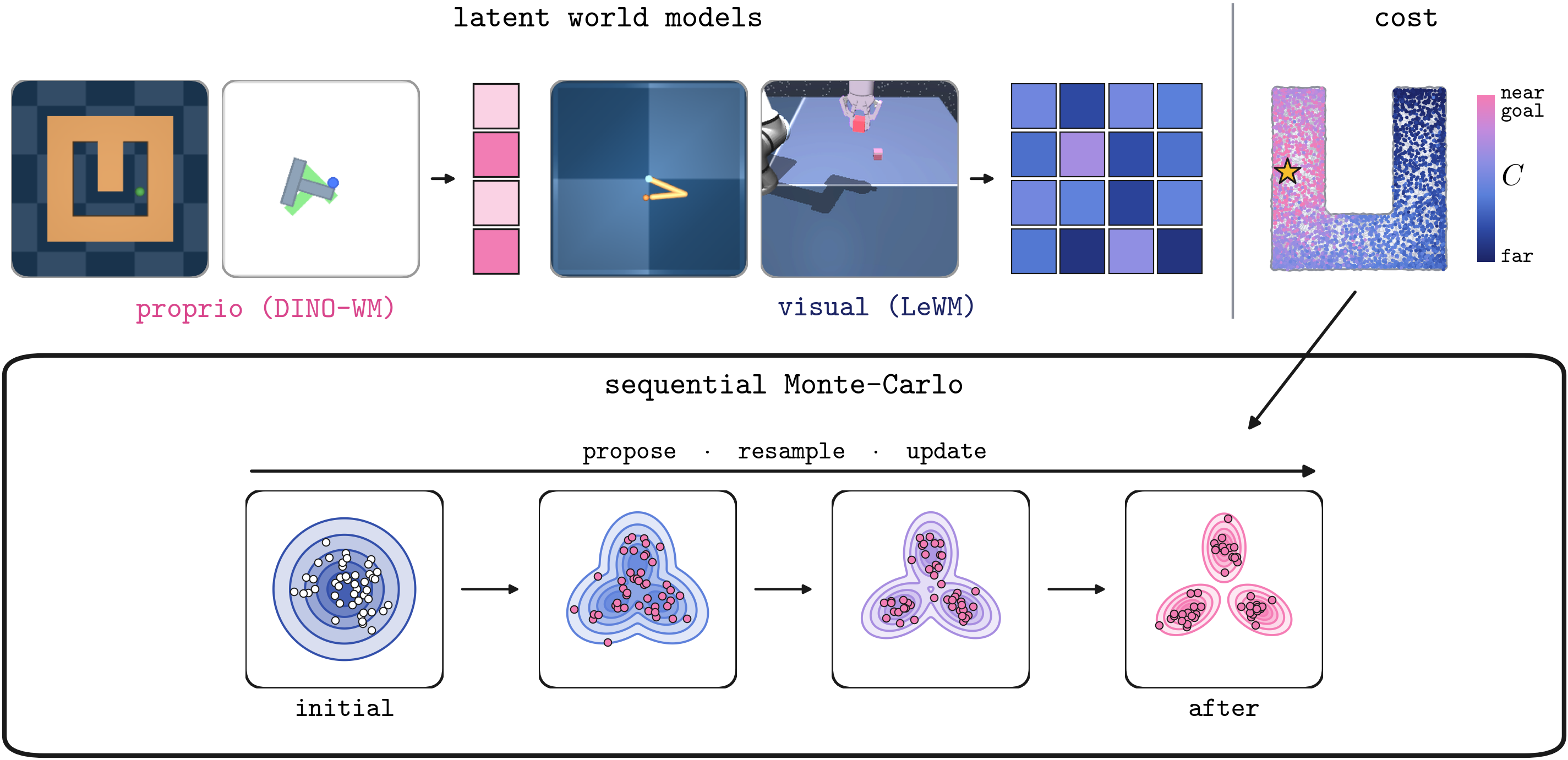}
    \caption{\textbf{Planning as sequential Monte Carlo over action sequences.} Top: latents of DINO-WM (proprioceptive) and LeWM (visual), and a Point maze cost-to-go (star: goal). Bottom (schematic): at each iteration, each particle (an action sequence) takes a Langevin step on the cost-defined target, the population is reweighted and, when its effective sample size is low, resampled, and an optimizer step moves the nominal sequence that indexes the target.}
    \label{fig:teaser}
\end{figure}

\section{Introduction}
World models learn compact representations of environments from observations and predict how they evolve under actions, enabling inexpensive simulation for planning \citep{zhou2025dinowmworldmodelspretrained, maes2026leworldmodelstableendtoendjointembedding, terver2025drives}. Planning with these models requires finding goal-reaching action sequences, typically by minimizing a cost that compares the predicted representations with the representation of a goal observation. Sampling-based planners, such as the cross-entropy method (CEM) \citep{zhou2025dinowmworldmodelspretrained, maes2026leworldmodelstableendtoendjointembedding}, draw action sequences from a Gaussian and refit it to the lowest-cost samples. While they require only cost evaluations, their performance degrades with longer planning horizons and higher action dimensionality \citep{psenka2026parallelstochasticgradientbasedplanning}. Gradient-based planners, on the other hand, differentiate the cost through the world model and perform well on tasks where the goal is greedily reachable \citep{terver2025drives}. However, they can become trapped in local minima \citep{terver2025drives}. Additional strategies have been proposed to overcome this, including optimizing the intermediate states alongside the actions with Langevin-style perturbations of these states, as in GRASP \citep{psenka2026parallelstochasticgradientbasedplanning}.

MPPI \citep{williams2016aggressive, williams2017model} takes a complementary approach, alternating between estimating an expectation under a cost-weighted distribution from these samples and updating the plan from this estimate. MPPI may thus be viewed as a form of expectation-maximization (EM), whose expectation step estimates the mean of the target from samples and whose maximization step moves the plan to this mean \cite{cuin2026learning, cuin2026efficientstochasticoptimisationsequential}. Its samples, however, are drawn afresh at every iteration from a Gaussian centered on the plan \citep[Alg.~1]{williams2016aggressive} and reach the target only through importance weights. The target distribution that favors goal-reaching action sequences is generally intractable to construct directly. Instead, it can be approached through a sequence of intermediate distributions, beginning with a Gaussian centered at the initial plan and gradually shifting toward lower-cost sequences as the plan is updated. Sequential Monte Carlo (SMC) provides a principled means of approximating such a sequence of distributions \citep{doucet2001introduction, del2004feynman}. SMC samplers \citep{del2006sequential}, extending annealed importance sampling \citep{neal1998annealedimportancesampling}, maintain a population of particles that is propagated between successive distributions using Markov kernels, with importance reweighting and resampling to maintain an approximation to each distribution \citep{arbel2021annealedflowtransportmonte, chen2025sequentialcontrolledlangevindiffusions}.

Our key insight is that latent-space planning can be formulated as optimization guided by SMC. The two methods complement each other, each carrying out one step of the EM view of MPPI. As the plan changes by one optimizer step per iteration, the targets of consecutive iterations form a sequence of evolving distributions. SMC carries a population of action sequences from one target to the next through sequential reweighting and resampling. The optimizer then updates the plan using this weighted population, thereby re-centering the target for the next iteration. This combines the population-based updates of MPPI with the gradient-based updates of conventional planners.

The tilting of action sequences toward the goal is determined by a cost function that assigns lower costs to more desirable sequences. A natural choice is the Euclidean distance between the final predicted latent and the goal latent, which is commonly used as an optimization objective in latent world-model planners \citep{zhou2025dinowmworldmodelspretrained, maes2026leworldmodelstableendtoendjointembedding}. Classical planning, however, often uses a cost-to-go, which quantifies the cost of reaching the goal from each state and can be obtained by solving a partial differential equation (PDE) over the feasible state space \citep{kimmel2001optimal, mirebeau2019hamiltonian}. Inspired by this, we introduce a PDE-based cost-to-go over the latent space to capture connectivity between states. Since latent world models do not provide explicit access to physical states, we construct a roadmap over latent states using transitions observed between physical states in the offline dataset and solve the PDE over this roadmap.

Our contributions are thus summarized as:
\begin{itemize}
    \item In \Cref{sec:latent_smc_planning}, we formulate latent-space planning as optimization guided by sequential Monte Carlo (SMC), combining population-based sampling with gradient-based plan updates through the world model.
    \item In \Cref{sec:cost}, we introduce a PDE-based cost-to-go over the latent space, inspired by classical planning. By accounting for connectivity between latent states through transitions in the offline dataset, this cost provides an alternative to the Euclidean distance commonly used in latent world-model planning.
    \item In \Cref{sec:experiments}, we evaluate our planner on navigation and manipulation tasks using DINO-WM and LeWM. Across twelve settings, our planner achieves a success rate of $79.8\%$, outperforming GD, GRASP, CEM, and MPPI by $28.8$, $49.8$, $34.5$, and $7.4$ percentage points, respectively, under a fixed budget of $150$ iterations. In seven of the twelve settings, we find that the PDE-based cost improves success over the $L_2$ cost and in the remaining five, the difference stays within $8$ percentage points.
\end{itemize}
\section{Background}
\label{sec:background}

\paragraph{Latent Planning}
World models such as DINO-WM \citep{zhou2025dinowmworldmodelspretrained} and LeWM \citep{maes2026leworldmodelstableendtoendjointembedding}, a joint-embedding predictive architecture (JEPA), define powerful latent spaces that facilitate planning entirely over the latents. These typically encode observations into a latent space and plan by minimizing the $L_2$ distance between the predicted final latent and the goal latent. They often use sampling-based planning such as CEM and MPPI. More recently, the differentiability of these models has enabled gradient-based optimization, with the recent GRASP \cite{psenka2026parallelstochasticgradientbasedplanning} offering efficient parallelized stochastic gradient-based planning. Other methods amortize or restructure the search with networks trained on the offline dataset, such as LeFlow \cite{huang2026leflow}, GC-IDM \cite{nguyen2026latent}, LEAP \cite{pham2026latent} and hierarchical world models \cite{zhang2026hierarchical}. In contrast, we formulate latent-space planning as an optimization procedure guided by SMC. Rather than using sampling or gradient-based optimization in isolation, our approach combines population-based exploration with gradient-based plan updates, while allowing the target distribution to evolve with the plan.

\paragraph{Sequential Monte Carlo.}
SMC has been applied to Bayesian inference, stochastic optimization, reinforcement learning, and trajectory planning \citep{hirt2019scalable, cuin2026efficientstochasticoptimisationsequential, abdulsamad2026sequential, piche2018probabilistic}. Recent work has also explored particle methods for diffusion sampling and reward alignment \citep{wu2026reverse, young2026diffusion, yoon2026psi, ou2026inferencetimescalingdiscretediffusion}, as well as learning latent-variable models \citep{cuin2026learning, wang2026traininglatentdiffusionmodels}. In trajectory planning, \citet{piche2018probabilistic} use SMC to extend partial trajectories one step at a time. In contrast, our particles represent complete action sequences, and each SMC iteration corresponds to a planner iteration. 

\paragraph{Planning over cost objectives.}
Classical planning often derives cost structure from environment geometry using PDEs \citep{kimmel2001optimal, mirebeau2019hamiltonian}. For example, minimum-time value functions satisfy the Eikonal equation under isotropic dynamics. This formulation has been used in neural motion planning \citep{ni2022ntfields, ni2023progressive} and reinforcement learning \citep{giammarino2026goal, giammarino2026physics, viswanath2026physics}, including as a physics-informed constraint \citep{raissi2019physics} for learning cost functions \citep{viswanath2026neural}. Another related approach is the heat method \citep{crane2012geodesicsheat}, which uses Varadhan's formula to approximate geodesic distances on grids, meshes, and point clouds. For sparsely covered or long-horizon settings, roadmap-based methods construct graphs over known states to capture environment topology. Such roadmaps have also been used to weakly supervise \citep{viswanath2026operator} continuous value functions through PDE-based formulations \citep{ ni2026weakly}. We investigate whether the same construction can provide a useful cost-to-go for latent-space planning, using a roadmap over encoded dataset states.

\section{Method}

We formulate latent-space planning as inference over distributions of action sequences. In the
path-integral control framework \cite{williams2016aggressive, williams2017model}, a cost function defines a target distribution that assigns greater probability to low-cost trajectories, and planning amounts to drawing
action sequences from this distribution. This formulation can be viewed as a form of EM that admits an iterative optimization procedure, alternating between estimating the
action distribution under the current cost and updating the distribution toward lower-cost trajectories. Recent works have proposed SMC as a framework for the E-step with the M-step instantiated
through first-order optimizers \cite{cuin2026learning, cuin2026efficientstochasticoptimisationsequential}. We follow a similar approach and formulate planning as SMC with optimization.

In this section, we start by setting up the problem. In \Cref{subsec:smc}, we recall how SMC carries a population of action sequences through a sequence of such targets via Markov kernels, reweighting and resampling it. Our planner (\Cref{sec:latent_smc_planning}) alternates between two steps (see also \Cref{alg:smc} and \Cref{fig:teaser}): an SMC step moves the population toward the current target, and an optimization step updates the plan from the weighted population and thereby re-centers the target. Finally, \Cref{sec:cost} defines the cost, including a PDE-derived cost-to-go on a roadmap of encoded dataset states.
\subsection{Problem Setup}
Let $\mathcal Z\subseteq\mathbb R^{d_z}$ be the latent space of a pretrained world model and $\mathcal A\subseteq\mathbb R^{m}$ the action space, with $\mathcal U\coloneqq\mathcal A^{H}$ the space of action sequences $\mathbf a=(a_0,\dots,a_{H-1})$ of horizon $H$. Given an initial latent $z_0$ and a goal $z_g$, we define a cost function $C:\mathcal U\to\mathbb R$ to score a sequence by its rollout $\mathcal F^{H}_{\theta_{\mathrm{WM}}}(z_0,\mathbf a)=(z_1,\dots,z_H)$ under the frozen world model $\mathcal{F}_{\theta_{\mathrm{WM}}}$. We define the target distribution
\begin{align}
\label{eq:target}
\pi_{\bm{\mu}}(\mathbf a)\propto q(\mathbf a\mid\bm{\mu})\exp\big(-C(\mathbf a)/\alpha\big),
\end{align}
where $q(\mathbf a\mid\bm{\mu})=\mathcal N(\bm{\mu},\sigma_p^2 \mathbf I)$ is a reference distribution centered on an \emph{action sequence} $\bm{\mu}\in\mathcal U$ and $\alpha>0$ is a temperature.

\subsection{Sequential Monte Carlo}\label{subsec:smc}
SMC approximates this target with $N$ weighted samples, called particles, which it carries over steps $k=0,\dots,K$ through the unnormalized targets $\gamma_k\coloneqq\gamma_{\bm{\mu}_k}$ of the plan-indexed family
\begin{equation}
\label{eq:target_family}
\gamma_{\bm{\mu}}(\mathbf a)= q(\mathbf a\mid\bm{\mu})\exp\!\big(-C(\mathbf a)/\alpha\big),
\qquad \pi_{\bm{\mu}}\propto\gamma_{\bm{\mu}},
\end{equation}
At each step $k$, SMC computes incremental importance weights using the next target, the forward Markov kernel $\mathsf M_k$, and a backward kernel $\mathsf L_{k-1}$. Particles are resampled when their weights become sufficiently concentrated, favoring particles with higher weights. The surviving particles are then propagated through the forward kernel, with their weights corrected for this transition using the backward kernel \citep{del2006sequential, arbel2021annealedflowtransportmonte, cuin2026efficientstochasticoptimisationsequential}. This construction defines a Feynman--Kac model \citep{del2004feynman, kim2025tuningsequentialmontecarlo}. We provide further background in \Cref{app:theory-fk}.

\subsection{Planning as Latent-Space SMC Optimization}
\label{sec:latent_smc_planning}

In the following, we define our planner, which maintains $N$ particles $\mathbf a_k^{(i)}\in\mathcal U$, each representing a complete action sequence, with associated weights $W_k^{(i)}$. Each iteration alternates between an SMC step, which moves the particles toward the next target, and an optimization step, which updates the nominal plan. Since the target depends on the nominal plan and the plan is updated from the weighted particles, the two steps form a coupled iteration. As noted by \citet{cuin2026learning}, establishing convergence guarantees for such iterations requires a nontrivial analysis, which we leave for future work. We describe the two steps below.

\paragraph{SMC step: population.} The initial particles are drawn from the reference, $\mathbf a_0^{(i)}\sim q(\cdot\mid\bm{\mu}_0)$, with the initial weight $W_0^{(i)}=\gamma_0(\mathbf a_0^{(i)})/q(\mathbf a_0^{(i)}\mid\bm{\mu}_0)=e^{-C(\mathbf a_0^{(i)})/\alpha}$, where $\alpha$ is the temperature. It is set to the standard deviation of the costs $C(\mathbf a_0^{(1)}),\dots,C(\mathbf a_0^{(N)})$. To carry the particles through the evolving targets $\gamma_k$, we use the Jarzynski-adjusted Langevin algorithm (JALA) of \citet{cuin2026learning}, whose main idea is to run the unadjusted Langevin algorithm (ULA) \citep{vempala2019rapid} on the time-varying potentials $U_k=-\log\gamma_k$ and to reweight the particles by Jarzynski's equality \citep[Prop.~1]{cuin2026learning}. At step $k$, each particle takes one ULA step at $\gamma_k$ with step size $\eta$,
\begin{equation}
\label{eq:move}
\mathbf a_k^{(i)}=\mathbf a_{k-1}^{(i)}+\eta\,\nabla_{\mathbf a}\log\gamma_k\big(\mathbf a_{k-1}^{(i)}\big)+\sqrt{2\eta}\,\varepsilon_k^{(i)},
\qquad\varepsilon_k^{(i)}\sim\mathcal N(0,\mathbf I),
\end{equation}
and its weight is multiplied by the incremental weight of an SMC sampler,
\begin{equation}
\label{eq:weights}
W_k^{(i)}=W_{k-1}^{(i)}\,G_k\big(\mathbf a_{k-1}^{(i)},\mathbf a_k^{(i)}\big),
\qquad
G_k(\mathbf x,\mathbf x')=\frac{\gamma_k(\mathbf x')\,\mathsf L_{k-1}(\mathbf x',\mathbf x)}{\gamma_{k-1}(\mathbf x)\,\mathsf M_k(\mathbf x,\mathbf x')},
\end{equation}
which is similar to Eq.~(3) of \citet{arbel2021annealedflowtransportmonte}, Eq.~(12) of \citet{del2006sequential} and Eq.~(8) of \citet{cuin2026efficientstochasticoptimisationsequential}, with $\mathsf M_k$ and $\mathsf L_{k-1}$ a representing the forward and backward kernels respectively. We normalize the weights of \Cref{eq:weights} and track their degeneracy by the effective sample size (ESS) \citep[Eq.~(12)]{cuin2026efficientstochasticoptimisationsequential},
\begin{equation}
\label{eq:ess}
w_k^{(i)}=\frac{W_k^{(i)}}{\sum_{j=1}^{N}W_k^{(j)}},\qquad
\operatorname{ESS}\big(W^{(1:N)}\big)=\frac{1}{\sum_{i}\big(w^{(i)}\big)^{2}}\,.
\end{equation}
When this ESS falls below a fraction $\tau$ of $N$, we resample the particles by systematic resampling. 

\paragraph{Optimization step: plan.} The optimization step updates the plan $\bm{\mu}$ to move the target toward low-cost sequences. Its objective is defined as 
$\ell(\bm{\mu})=-\log\int_{\mathcal U}\gamma_{\bm{\mu}}(\mathbf a)\,\mathrm d\mathbf a=-\log\E_{q(\cdot\mid\bm{\mu})}\big[e^{-C(\mathbf a)/\alpha}\big]$, a smoothed minimum of $C/\alpha$ over the reference. By Fisher's identity \citep[Eq.~(5)]{cuin2026learning}, the gradient is an expectation under the target distribution. We approximate this expectation using the weighted particles and pass the resulting estimate, with \(U_{\bm{\mu}}=-\log\gamma_{\bm{\mu}}\), to the optimizer \textsc{opt} (Adam):
\begin{equation}
\label{eq:mstep}
\mathbf g_k=\textstyle\sum_{i=1}^{N}w_k^{(i)}\,\nabla_{\bm{\mu}}U_{\bm{\mu}_k}\big(\mathbf a_k^{(i)}\big)
=-\frac{1}{\sigma_p^{2}}\sum_{i=1}^{N}w_k^{(i)}\big(\mathbf a_k^{(i)}-\bm{\mu}_k\big),
\qquad
\bm{\mu}_{k+1}=\textsc{opt}(\bm{\mu}_k,\mathbf g_k).
\end{equation}
This is analogous to the parameter update of SMC for stochastic optimization \citep{cuin2026learning, cuin2026efficientstochasticoptimisationsequential}, but \textit{crucially}, we reformulate it for planning: the quantity it optimizes is the plan itself, as opposed to the model parameter. 

\begin{algorithm}[t]
\caption{Latent-space planning as an SMC sampler with an optimization step on the nominal sequence; symbols as in \Cref{sec:latent_smc_planning}. Lines with superscript $(i)$ run for $i=1,\dots,N$. }
\label{alg:smc}
\begin{algorithmic}[1]
\Require initial latent $z_0$, goal latent $z_g$, frozen world model $\mathcal F_{\theta_{\mathrm{WM}}}$, cost $C$, nominal sequence $\bm{\mu}_0=\mathbf 0$, reference width $\sigma_p$, particles $N$, iterations $K$, step size $\eta$, ESS fraction $\tau$, optimizer \textsc{opt}
\State $\mathbf a_0^{(i)}\sim q(\cdot\mid\bm{\mu}_0)$;\; $\bm{\mu}_1\leftarrow\bm{\mu}_0$;\; $\lambda_V$ as in \Cref{sec:cost};\; $\alpha\leftarrow\operatorname{std}_i C(\mathbf a_0^{(i)})$;\; $W_0^{(i)}\leftarrow e^{-C(\mathbf a_0^{(i)})/\alpha}$
\For{$k=1,\dots,K$}
  \State $\widetilde W_k^{(i)}\leftarrow W_{k-1}^{(i)}\,\gamma_k\big(\mathbf a_{k-1}^{(i)}\big)\big/\gamma_{k-1}\big(\mathbf a_{k-1}^{(i)}\big)$ \Comment{change of target; part of $G_k$ (\plaineqref{eq:weights})}
  \If{$\operatorname{ESS}\big(\widetilde W_k^{(1:N)}\big)<\tau N$} \Comment{(\plaineqref{eq:ess})}
    \State draw $I_k^{(1:N)}$ by systematic resampling with probabilities $\propto\widetilde W_k^{(1:N)}$
    \State $\mathbf a_{k-1}^{(i)}\leftarrow\mathbf a_{k-1}^{(I_k^{(i)})}$;\; $\widetilde W_k^{(i)}\leftarrow 1$
  \EndIf
  \State $\mathbf a_k^{(i)}\leftarrow$ ULA move (\plaineqref{eq:move}) from $\mathbf a_{k-1}^{(i)}$ \Comment{SMC step: no accept--reject}
  \State $W_k^{(i)}\leftarrow\widetilde W_k^{(i)}\,\gamma_k\big(\mathbf a_k^{(i)}\big)\mathsf M_k\big(\mathbf a_k^{(i)},\mathbf a_{k-1}^{(i)}\big)\big/\big[\gamma_k\big(\mathbf a_{k-1}^{(i)}\big)\mathsf M_k\big(\mathbf a_{k-1}^{(i)},\mathbf a_k^{(i)}\big)\big]$ \Comment{Jarzynski correction; completes $G_k$}
  \State $w_k^{(i)}\leftarrow W_k^{(i)}/\sum_{j}W_k^{(j)}$
  \State $\mathbf g_k\leftarrow-\sigma_p^{-2}\sum_{i=1}^{N}w_k^{(i)}\big(\mathbf a_k^{(i)}-\bm{\mu}_k\big)$ \Comment{optimization-step gradient (\plaineqref{eq:mstep}; \Cref{app:em})}
  \State $\bm{\mu}_{k+1}\leftarrow\textsc{opt}(\bm{\mu}_k,\mathbf g_k)$ \Comment{optimization step: updates $\bm{\mu}$, not the particles}
\EndFor
\State \Return the particles $\{\mathbf a_k^{(i)}\}_{1\le k\le K,\,i}$ and the plans $\{\bm{\mu}_k\}_{2\le k\le K+1}$ \Comment{one is executed after each iteration}
\end{algorithmic}
\end{algorithm}

\subsection{Defining Cost in Latent Spaces}
\label{sec:cost}

In classical planning, the cost-to-go is obtained by solving a PDE over the feasible state space \citep{kimmel2001optimal, mirebeau2019hamiltonian}. Unlike Euclidean distance, it reflects the environment's geometry: paths to the goal must respect obstacles and other constraints. Since a latent world model does not explicitly represent the feasible state space, we construct a \textit{roadmap} over dataset latents, with edges defined by observed transitions, to approximate its connectivity.

\begin{definition}[Latent Roadmap]
\label{def}
Let $E_{\theta_{\mathrm{WM}}}$ denote the encoder of the frozen world model. We construct a graph $\mathcal{R}=(\mathcal{V},\mathcal{E})$ over the offline dataset, with vertices
$\mathcal{V}=\{\xi_i\}_{i=1}^{N_{\mathcal R}}$,
where $\xi_i=E_{\theta_{\mathrm{WM}}}(x_i)$ is the latent encoding of a dataset state $x_i$. For each observed transition from $x_i$ to $x_i'$, let $\xi_i'=E_{\theta_{\mathrm{WM}}}(x_i')$, and let $\mathcal{N}_i$ contain the $\kappa$ vertices nearest to $\xi_i'$. We add an edge $(\xi_i,\xi_j)$ for each $j\in\mathcal{N}_i\setminus\{i\}$.
\end{definition}

We assign these edges affinities
\[
\tilde\omega_{ij} =
\begin{cases}
\exp\!\left(-\lVert\xi_i'-\xi_j\rVert^2/(2\sigma_i^2)\right),
& j\in\mathcal{N}_i\setminus\{i\},\\
0, & \text{otherwise},
\end{cases}
\]
where $\sigma_i$ is twice the second-smallest distance from $\xi_i'$ to a vertex in $\mathcal{N}_i$. We symmetrize the affinity matrix,
$\Omega=(\widetilde\Omega+\widetilde\Omega^\top)/2$,
and define the degree matrix by $D_{ii}=\sum_j\omega_{ij}$. The resulting unnormalized graph Laplacian is $L=D-\Omega$.

Following the heat method of \citet{crane2012geodesicsheat}, we compute a field on the roadmap by taking one backward Euler step of the heat equation, initialized at the vertex nearest the goal latent:
\begin{equation}
g=\argmin_i\lVert\xi_i-z_g\rVert,\qquad
(I+tL)u=e_g,
\label{eq:heat}
\end{equation}
where $e_g$ is the unit vector at vertex $g$ and $t>0$ is the time step. Here, $-L$ plays the role of the discrete Laplace--Beltrami operator on the roadmap. The resulting field $u$ is largest at the goal. Although the heat method relates this field to geodesic distance through the Varadhan estimate $\sqrt{-4t\log u}$, this estimate is sensitive to errors in the magnitude of $u$ \citep{crane2012geodesicsheat}. We therefore use only the ordering induced by $u$ and define the cost-to-go at vertex $i$ by its normalized rank:
\begin{equation}
V_i=
\frac{\operatorname{rank}_i(-\log u)-1}{N_{\mathcal R}-1},
\label{eq:rank}
\end{equation}
where tied values receive their average rank. Thus, $V_g=0$ and the cost increases as $u$ decreases. Its absolute scale is immaterial because $\lambda_V$ controls the contribution of this term to the planning cost. We set $t=c/\operatorname{median}_i D_{ii}$, with $c$ and $\kappa$ selected separately for each environment using only the offline dataset (\Cref{app:impl}). The effect of $t$ and alternative constructions are examined in \Cref{app:operator}.

For a predicted action sequence $\mathbf a$, we evaluate the cost-to-go at each predicted latent $z_s$ and take a weighted average:
\begin{equation}
C_V(\mathbf a)=\sum_{s=1}^{H}\nu_s V(z_s),
\qquad
\nu_s=\frac{2^{s-1}}{2^H-1}.
\label{eq:pde-cost}
\end{equation}
The weights increase along the horizon, giving greater weight to later states while retaining a contribution from every step. The full planning cost is
\begin{equation}
C(\mathbf a)=\mathcal L(\mathbf a)
+\lambda_a\,\frac{1}{Hm}\lVert\mathbf a\rVert^2
+\lambda_V C_V(\mathbf a),
\label{eq:cost}
\end{equation}
where $\mathcal L(\mathbf a)$ is the mean squared distance between the final predicted latent and the goal latent, summed over visual and proprioceptive components when both are available. The quadratic term penalizes the standardized actions, following path-integral control \citep[Eq.~(2)]{williams2016aggressive}. Dividing the term by $Hm$ makes its scale independent of the horizon. In the experiments, the \emph{PDE cost} sets $\lambda_V$ to the ratio of the standard deviations of $\mathcal L$ and $C_V$ over the initial population, and the \emph{$L_2$ cost} uses the $L_2$ objective. 

\section{Experiments}
\label{sec:experiments}
In this section, we discuss our experiments designed to answer the following questions: (Q1) how our planner compares with zero-order and gradient-based planners at the same number of particles and iterations (\Cref{sec:exp:planning}), (Q2) whether our planner and each baseline solve the same seeds (\Cref{sec:exp:horizon}), and (Q3) whether the PDE cost improves success over the $L_2$ cost (\Cref{sec:exp:ablation}).

\subsection{Setup}

\paragraph{Environments, world models and baselines.}
We evaluate on five environments: Reacher, Point maze, Wall, Push-T and Cube. Reacher and Cube are evaluated on LeWM, which encodes each frame as a single $192$-dimensional token and exposes no proprioceptive latent, so the roadmap of \Cref{sec:cost} is built on this token. Point maze, Wall and Push-T are solved using DINO-WM, which encodes each frame as $196$ patch tokens and also exposes a $10$-dimensional proprioceptive latent, on which the roadmap is built. 

We compare against four planners: CEM, MPPI \citep{williams2016aggressive, williams2017model}, gradient descent (GD) and GRASP \citep{psenka2026parallelstochasticgradientbasedplanning}. 
In \Cref{tab:planners}, each baseline optimizes the objective reported by its authors, and \Cref{app:abl-cost} examines the effect of our PDE cost on GD, CEM and MPPI. We present our hyperparameter details in \Cref{tab:hyper} and baseline configurations in \Cref{app:baselines}. We further investigate the impact of SMC kernels by considering the Metropolis-adjusted Langevin algorithm (MALA), random-walk Metropolis (RWM) and Hamiltonian Monte Carlo (HMC). We discuss our findings in \Cref{app:kernel-abl}. 

\paragraph{Budget.}
We evaluate all planners under a matched search budget, by which we mean the number of candidate action trajectories whose cost is evaluated in parallel and the number of iterations executed per plan. We cap the number of iterations at 150 for every planner and set the number of candidates to 12. In our case, that corresponds to the population of the particles. We run all planners over the same pretrained world model, across the same seeds with the same start and goal pairs. We measure the planning time to success by the number of iterations to the first reported success. \Cref{tab:planners} compares the planners at this small, equal number of particles. We do not compare wall-clock time or memory, which depend on how each implementation parallelizes its computation and on whether it backpropagates through the world model, as GD, GRASP and our planner do (\Cref{app:baselines}). We additionally evaluate GD and GRASP with their released configurations and CEM at an equal rollout budget to a single-particle gradient planner (\Cref{app:abl-baselines}).

\paragraph{Evaluation.}
For every environment except Cube, we consider $100$ trials (seeds), and $50$ on Cube, each with its own start and goal. A trial is a single open-loop planning problem, as in \citet{zhou2025dinowmworldmodelspretrained} wherein the planner optimizes one sequence of $H$ world-model actions from the start observation toward the goal observation. We measure the planning horizon $H$ in world-model steps, the number of latent transitions predicted by the world model, and evaluate $H\in\{5,10,15\}$ ($H{=}5$ on Cube). Each world-model action is a block of $5$ environment actions, the frameskip of \citet{zhou2025dinowmworldmodelspretrained} and \citet{maes2026leworldmodelstableendtoendjointembedding}, so a plan spans $5H$ environment steps. 

After every iteration, the current plan of each planner is executed in the simulator from the start state, and the trial succeeds as soon as the state reached at the end of the plan meets the success criterion of the environment. Planning stops at the first success. The number of iterations to success is the index of the first iteration whose plan succeeds, and a trial that does not succeed within $150$ iterations is a failure.

\subsection{Quantitative Results}
\label{sec:exp:planning}
\begin{figure}[!h]
  \centering
  \includegraphics[width=\linewidth]{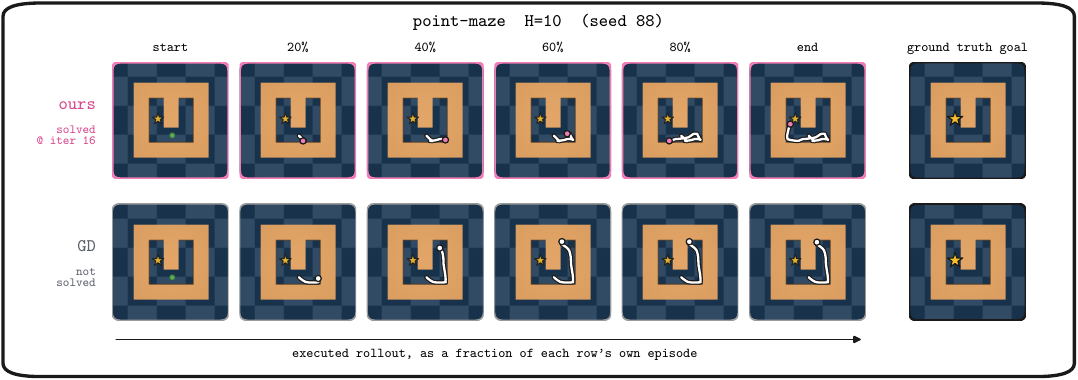}
  \caption{Point maze rollouts ($H{=}10$, seed 88) of our planner with $12$ particles and of GD with $12$ candidates. The run shown uses the estimator of \Cref{eq:mstep-path} and solves in $16$ iterations; the planner of \Cref{tab:planners}, which on Point maze uses $\mathbf g_k$ of \Cref{eq:mstep}, solves this seed in $3$, and GD reaches the $150$-iteration cap.}
  \label{fig:frames_pm}
\end{figure}

\begin{figure}[!h]
  \centering
  \includegraphics[width=\linewidth]{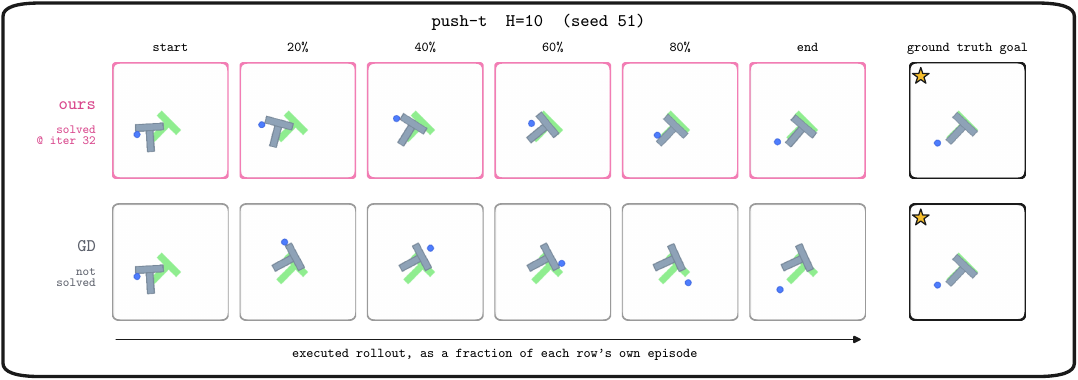}
  \caption{Push-T rollouts ($H{=}10$, seed 51) of our planner with $12$ particles and of GD with $12$ candidates, on a seed whose goal places the T on the green target drawn by the environment. Ours solves in $32$ iterations; GD, GRASP, CEM and MPPI do not solve this seed within $150$ iterations.}
  \label{fig:frames_pusht}
\end{figure}

We begin by comparing the success rates of our planner against the baselines across all tasks and planning horizons, as summarized in \Cref{tab:planners}. Overall, our planner achieves the highest success rate in eight of the twelve task–horizon settings, demonstrating consistent performance across the evaluated environments. To assess whether these improvements are statistically meaningful, we conduct paired comparisons over the evaluation seeds. Our planner significantly outperforms the strongest baseline in four settings (exact McNemar test at the $5\%$ significance level, with Holm correction across the twelve settings; \Cref{app:ci}), namely Reacher at all three horizons and Wall at $H{=}5$. In the remaining four settings, a baseline achieves a higher success rate, with differences ranging from 1 to 6 percentage points. MPPI, which is the strongest baseline in ten settings, leads on Point maze at $H{=}15$ and Wall at $H{=}10$ and $H{=}15$, while GD performs best on Push-T at $H{=}15$. We illustrate the resulting planning behavior through representative rollouts on Point maze (\Cref{fig:frames_pm}) and Push-T (\Cref{fig:frames_pusht}).

\definecolor{bestpurple}{HTML}{5B2C87}
\begin{table}[t]
  \caption{Success rate (\%), with the mean iterations to success of successful seeds in parentheses, at horizons $H\in\{5,10,15\}$; $12$ particles, $150$ iterations, $100$ seeds per cell. Bold: best per column.}
  \label{tab:planners}
  \centering
  \setlength{\tabcolsep}{7pt}
  \renewcommand{\arraystretch}{1.15}
  \resizebox{1\textwidth}{!}{%
  \begin{tabular}{lcccccccccccc}
    \toprule
    & \multicolumn{3}{c}{Reacher} & \multicolumn{3}{c}{Point maze} & \multicolumn{3}{c}{Wall} & \multicolumn{3}{c}{Push-T} \\
    \cmidrule(lr){2-4} \cmidrule(lr){5-7} \cmidrule(lr){8-10} \cmidrule(lr){11-13}
    Planner & 5 & 10 & 15 & 5 & 10 & 15 & 5 & 10 & 15 & 5 & 10 & 15 \\
    \midrule
    GD & 83 \tiny{(24)} & 38 \tiny{(49)} & 15 \tiny{(48)} & 49 \tiny{(29)} & 66 \tiny{(18)} & 69 \tiny{(22)} & 58 \tiny{(32)} & 58 \tiny{(41)} & 59 \tiny{(43)} & 87 \tiny{(34)} & 24 \tiny{(50)} & \best{5} \tiny{(90)} \\
    GRASP & 36 \tiny{(76)} & 47 \tiny{(62)} & 59 \tiny{(58)} & 40 \tiny{(55)} & 49 \tiny{(37)} & 63 \tiny{(35)} & 10 \tiny{(102)} & 10 \tiny{(108)} & 4 \tiny{(116)} & 36 \tiny{(107)} & 5 \tiny{(103)} & 0 \\
    CEM & 69 \tiny{(4)} & 42 \tiny{(5)} & 41 \tiny{(5)} & 59 \tiny{(4)} & 75 \tiny{(5)} & 80 \tiny{(5)} & 29 \tiny{(9)} & 41 \tiny{(7)} & 56 \tiny{(8)} & 42 \tiny{(7)} & 6 \tiny{(8)} & 3 \tiny{(9)} \\
    MPPI & 86 \tiny{(11)} & 78 \tiny{(15)} & 67 \tiny{(18)} & 82 \tiny{(19)} & 95 \tiny{(13)} & \best{97} \tiny{(9)} & 78 \tiny{(34)} & 87 \tiny{(30)} & 86 \tiny{(29)} & 83 \tiny{(32)} & 27 \tiny{(48)} & 2 \tiny{(43)} \\
    \midrule
    \textbf{Ours (PDE cost)} & \best{100} \tiny{(10)} & \best{100} \tiny{(25)} & \best{89} \tiny{(33)} & \best{88} \tiny{(16)} & \best{98} \tiny{(9)} & 96 \tiny{(9)} & \best{92} \tiny{(23)} & 81 \tiny{(19)} & 84 \tiny{(19)} & 93 \tiny{(20)} & 32 \tiny{(53)} & 4 \tiny{(73)} \\
    \textbf{Ours ($L_2$ cost)} & 99 \tiny{(12)} & 78 \tiny{(25)} & 27 \tiny{(33)} & 79 \tiny{(9)} & 91 \tiny{(8)} & 95 \tiny{(12)} & 89 \tiny{(26)} & \best{89} \tiny{(23)} & \best{91} \tiny{(26)} & \best{94} \tiny{(26)} & \best{35} \tiny{(54)} & 4 \tiny{(87)} \\
    \bottomrule
  \end{tabular}}
\end{table}

\begin{figure}[!htbp]
  \centering
  \includegraphics[width=0.315\textwidth]{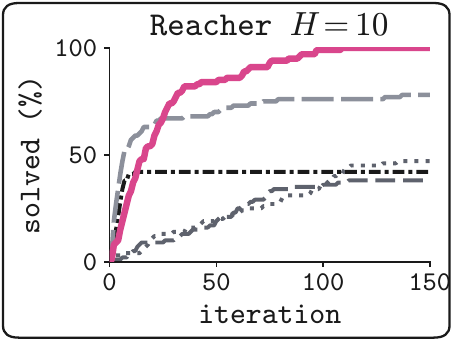} \hfill
  \includegraphics[width=0.315\textwidth]{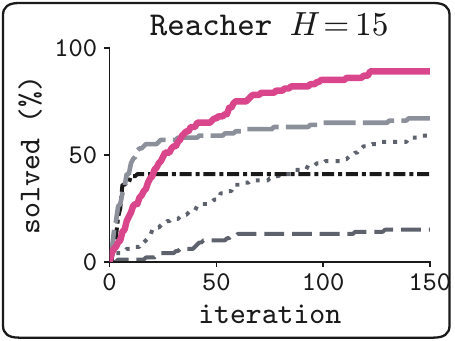} \hfill
  \includegraphics[width=0.315\textwidth]{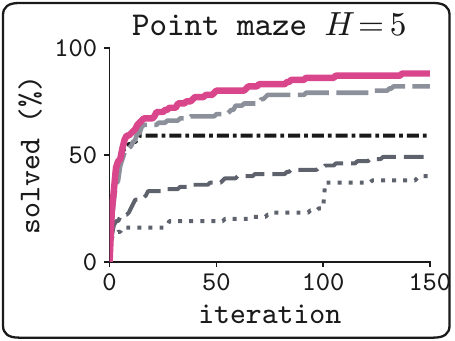} \\[4pt]
  \includegraphics[width=0.315\textwidth]{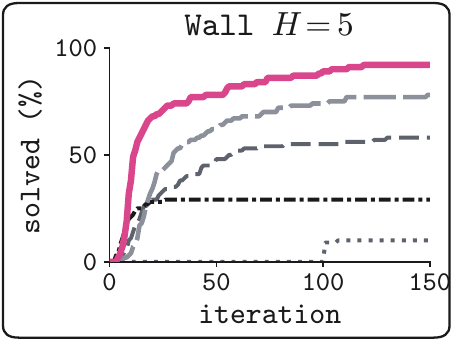} \hfill
  \includegraphics[width=0.315\textwidth]{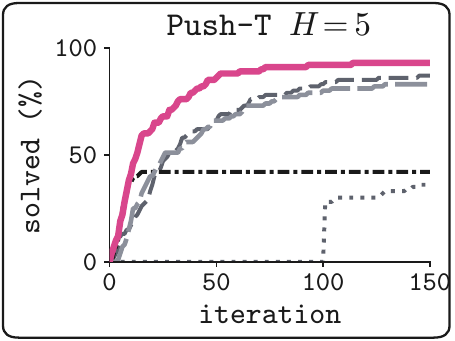} \hfill
  \includegraphics[width=0.315\textwidth]{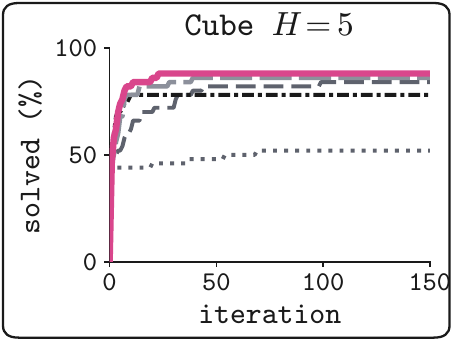}
  \\[2pt]
  \includegraphics[width=0.62\textwidth]{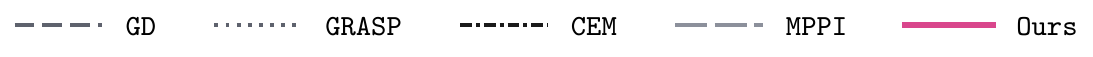}
  \caption{Fraction of seeds solved at or before each iteration, with success measured by executing each plan in the simulator and every planner holding $12$ particles. Each environment is shown at the horizon at which the margin of our planner over the strongest baseline is largest, Reacher at both $H{=}10$ and $H{=}15$, where the margins are equal. \Cref{app:curves} shows the other settings. Right-hand endpoints are the success rates of \Cref{tab:planners} and, for Cube, of \Cref{sec:exp:planning}.}
  \label{fig:success}
\end{figure}

We further observe that successful runs of our planner take fewer iterations on average than those of GD in eleven of the twelve settings, of GRASP in all eleven in which GRASP succeeds, and of MPPI in seven, all but Reacher and Push-T at $H{=}10$ and $H{=}15$, with a tie on Point maze at $H{=}15$. Successful runs of CEM take $4$ to $9$ iterations on average, but its success rate saturates quickly, whereas our planner keeps solving further seeds over more iterations and ends with higher success than CEM in every setting. In the eight settings in which our planner ends ahead, it exceeds the $150$-iteration success rate of the strongest baseline after $21$ to $104$ iterations (\Cref{fig:success,fig:curves_rest}).

\paragraph{Cube.} We evaluate OGBench Cube \citep{park2025ogbench} at $H{=}5$, with goals $25$ environment steps ahead in held-out expert episodes. On $21$ of the $50$ seeds the block already lies within $0.04$\,m of its target at the start; every planner solves these seeds, and we also report success on the other $29$. Our planner reaches $88\%$ on all seeds and $79\%$ on the other $29$, against $86\%$ and $76\%$ for MPPI, $84\%$ and $72\%$ for GD, $78\%$ and $62\%$ for CEM, and $52\%$ and $17\%$ for GRASP (\Cref{app:ci}).

\paragraph{Planners that train additional networks.} In \Cref{tab:learned}, we compare our planner with GC-IDM \cite{nguyen2026latent} and with LeFlow \cite{huang2026leflow}. We evaluate LeFlow under two settings: with $12$ sample paths, matching the number of particles of \Cref{tab:planners}, and with $64$, the default setting for the planner. Each planner runs with its released code on the LeWM environments (\Cref{app:baselines}). Our planner succeeds more often than both in every setting ($p<0.01$).

We provide ablation results and additional qualitative evaluations in \Cref{app:ablations}.

\begin{table}[t]
  \caption{Success rate (\%) with $95\%$ Wilson interval of planners that train additional networks and of ours$^{\ast}$ (\Cref{tab:hyper}), on $100$ seeds per Reacher horizon and $50$ on Cube; $p$: exact McNemar test of ours against each row.}
  \label{tab:learned}
  \centering
  \scriptsize
  \setlength{\tabcolsep}{2.5pt}
  \begin{tabular}{lcccccccc}
    \toprule
    & \multicolumn{2}{c}{Reacher $H{=}5$} & \multicolumn{2}{c}{Reacher $H{=}10$} & \multicolumn{2}{c}{Reacher $H{=}15$} & \multicolumn{2}{c}{Cube $H{=}5$} \\
    \cmidrule(lr){2-3}\cmidrule(lr){4-5}\cmidrule(lr){6-7}\cmidrule(lr){8-9}
    Planner & Success (\%) & $p$ & Success (\%) & $p$ & Success (\%) & $p$ & Success (\%) & $p$ \\
    \midrule
    Ours$^{\ast}$ & 100 [96, 100] & -- & 100 [96, 100] & -- & 89 [81, 94] & -- & 88 [76, 94] & -- \\
    \midrule
    LeFlow, $12$ paths & 83 [74, 89] & $<$0.001 & 64 [54, 73] & $<$0.001 & 29 [21, 39] & $<$0.001 & 58 [44, 71] & $<$0.001 \\
    LeFlow, $64$ paths & 89 [81, 94] & $<$0.001 & 66 [56, 75] & $<$0.001 & 40 [31, 50] & $<$0.001 & 62 [48, 74] & 0.002 \\
    GC-IDM & 82 [73, 88] & $<$0.001 & 59 [49, 68] & $<$0.001 & 35 [26, 45] & $<$0.001 & 56 [42, 69] & $<$0.001 \\
    \bottomrule
  \end{tabular}
\end{table}

\subsection{Paired Seed Outcomes and Failure Modes}\label{sec:exp:horizon}

\begin{figure}[t]
  \centering
  \includegraphics[width=\linewidth]{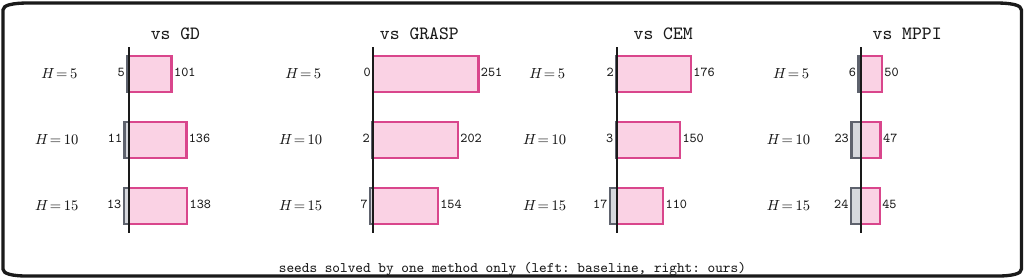}
  \caption{Paired outcomes against each baseline of \Cref{tab:planners}, every planner with $12$ particles: one row per horizon, each summed over the $400$ seeds of the four environments. Bars to the right count seeds our planner solves and the baseline does not, and bars to the left count the reverse. Per-environment panels are in \Cref{app:paired}.}
  \label{fig:paired_summary}
\end{figure}

We investigate whether the models fail on the same seeds through paired-seed tests. Two planners with the same success rate may solve entirely different seeds. We observe that across the twelve settings, our planner solves 375 seeds that GD does not, while GD solves 29 that ours does not. The corresponding counts are 607/9 for GRASP, 436/22 for CEM, and 142/53 for MPPI (\Cref{fig:paired_summary}). Most baseline-only successes occur at longer horizons: 24 of 29 for GD, all 9 for GRASP, 20 of 22 for CEM, and 47 of 53 for MPPI. These 113 outcomes correspond to 69 distinct seeds, 60 at $H{=}10$ or $H{=}15$.

The failures fall into two groups. On Wall at $H{=}10$ and $H{=}15$ and on Reacher at $H{=}15$, the executed plans often achieve low latent goal distance but fail to meet the simulator success criterion. Specifically, 19 of 26 such Wall seeds and all 11 Reacher seeds reach below the median latent goal distance of successful runs in the same setting. Yet on Wall, the median final position errors are $6.1$ and $7.1$ at $H{=}10$ and $H{=}15$, respectively, exceeding the success threshold of $4.5$. This suggests a mismatch between latent goal distance and the actual reached state, rather than insufficient search. Push-T exhibits the opposite failure: only 7 of its 21 baseline-only seeds reach the corresponding latent goal-distance level within 150 iterations, suggesting insufficient optimization. Neither failure is explained by weight collapse: the median ESS over the final 20 iterations is $0.56N$, and falls below $0.3N$ on only 2 of the 69 seeds.

\begin{figure}[!h]
  \centering
  \includegraphics[width=\linewidth]{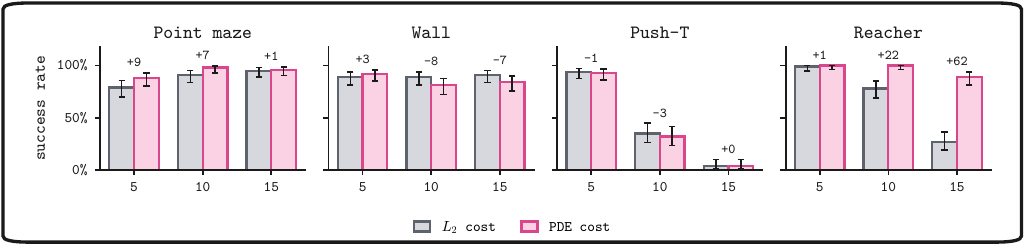}
  \caption{Success of our planner of \Cref{tab:planners} (PDE cost) and of the same planner with the $L_2$ cost ($\lambda_V=0$) on the same $100$ seeds per setting, by environment and horizon, with the difference in points above each pair. Error bars are $95\%$ Wilson intervals for each rate separately; significance is the paired McNemar test of \Cref{app:ci}, not the overlap of these intervals.}
  \label{fig:ablation}
\end{figure}

\subsection{Impact of the PDE Cost}
\label{sec:exp:ablation}

Lastly, we investigate the contribution of the PDE-based cost structure introduced in \Cref{sec:cost}. \Cref{fig:ablation} compares the success rates of the PDE and $L_2$ costs under an identical optimization budget. The largest differences arise on Reacher at $H{=}10$ and $H{=}15$, where the PDE cost improves success by 22 and 62 percentage points, respectively. More specifically, it solves 22 and 65 seeds that the $L_2$ cost fails to solve, compared with 0 and 3 seeds in the reverse direction (exact McNemar test on paired seeds, $p<0.001$ for both settings; \Cref{app:ci}). On Point maze, the PDE cost achieves higher success rates by 9, 7, and 1 percentage points at $H{=}5$, $H{=}10$, and $H{=}15$, respectively, although the differences at the first two horizons are significant only before Holm correction ($p=0.049$ and $p=0.016$). Conversely, the $L_2$ cost leads on Wall by 8 and 7 percentage points at $H{=}10$ and $H{=}15$, respectively, but neither difference is statistically significant ($p=0.12$ and $p=0.092$). In the remaining settings, the two costs yield comparable success rates, differing by at most 3 percentage points. This also holds on Cube, where they solve 44 and 43 of the 50 seeds, respectively.

To further examine how the cost structures reflect the underlying task geometry, we compare the offline rank correlation of $V$ with the observed number of steps to the goal against that of Euclidean distance in the roadmap latent space. The PDE-based cost exhibits stronger correlation on Point maze and Reacher, but not on Push-T (\Cref{tab:calib-operators}; \Cref{app:pde-scope}). We believe that this difference stems from the nature of the tasks: Point maze and Reacher involve more constrained motions, whereas Push-T permits relatively unconstrained movement through free space. Finally, \Cref{app:abl-mstep} examines the two cost structures under each form of the optimization step.

\section{Conclusion}

We formulated planning with learned world models as sequential Monte Carlo over complete action sequences in latent space. The planner maintains a population of candidates, using importance weighting, resampling, and stochastic propagation to guide the search toward goal-reaching trajectories. Across five environments and two learned world models, it improves success over zero-order, gradient-based and stochastic gradient-based planners, and over MPPI, the strongest of them in ten of the twelve settings.

We also investigated how the cost defining the target distribution affects planning. Beyond the conventional $L_2$ objective, we introduced a PDE-based cost-to-go defined over a roadmap of offline transitions. By incorporating local connectivity, this cost captures task-relevant structure beyond latent-space distance alone. Its effectiveness, however, depends on the coverage and representational capacity of the roadmap. Establishing convergence guarantees for the coupled SMC and optimization updates remains an open problem. Overall, our results establish population-based inference as an effective approach to planning with
learned world models.

\clearpage
\section*{Disclosure of Generative AI Use}
Generative AI tools (ChatGPT and Claude) assisted with manuscript drafting and revision, mathematical exposition and proof writing, interpretation of existing results, and LaTeX formatting, figure formatting and consistency checks. The authors are responsible for the final manuscript, including its mathematical claims, empirical reporting, and AI-assisted content.
\clearpage

\bibliography{iclr2027_conference}
\bibliographystyle{iclr2027_conference}

\appendix

\clearpage
\crefalias{section}{appendixsec}
\crefalias{subsection}{appendixsec}
\crefalias{subsubsection}{appendixsec}

\section{Theoretical background}
\label{app:theory}

\subsection{Target distribution}
\label{app:theory-target}

\begin{proposition}[{KL-regularized target; the Gibbs variational principle, as stated by \citet{yoon2026psi}}]
\label{prop:kl-target}
Let $p_0$ be a reference distribution, $r$ a reward and $\alpha\in\mathbb R_+$ the strength of the regularization. The distribution
\begin{equation}
\label{eq:app-kl}
p_0^{*}=\argmax_{q}\ \E_{\mathbf x_0\sim q}\big[r(\mathbf x_0)\big]-\alpha\,\KL\big(q\,\|\,p_0\big)
\end{equation}
has the closed form $p_0^{*}(\mathbf x_0)=\frac{1}{Z_0}\,p_0(\mathbf x_0)\exp\big(r(\mathbf x_0)/\alpha\big)$, where $Z_0$ is the normalizing constant.
\end{proposition}


\subsection{Feynman--Kac models and SMC samplers}
\label{app:theory-fk}

\begin{definition}[{Feynman--Kac model; from \citet{kim2025tuningsequentialmontecarlo}}]
\label{def:fk}
Let $\mathcal X$ be a space with a $\sigma$-finite base measure, $q$ the reference or initial proposal distribution, $(\mathsf M_k)_{k\in[K]}$ Markov kernels and $G_0:\mathcal X\to\mathbb R_{\ge0}$ and $(G_k)_{k\in[K]}$ non-negative measurable functions on $\mathcal X^2$, called potentials, where $[K]\coloneqq\{1,\dots,K\}$. The Feynman--Kac model describes a change of measure between the target and proposal path distributions
\begin{equation}
\label{eq:app-fk}
\begin{aligned}
P_{0:K}(\mathrm dx_{0:K})&\coloneqq\frac{1}{Z_K}\Big\{G_0(x_0)\prod_{k=1}^{K}G_k(x_{k-1},x_k)\Big\}Q_{0:K}(\mathrm dx_{0:K}),\\
Q_{0:K}(\mathrm dx_{0:K})&\coloneqq q(\mathrm dx_0)\prod_{k=1}^{K}\mathsf M_k(x_{k-1},\mathrm dx_k),
\end{aligned}
\end{equation}
with intermediate normalizing constants $Z_k=\int_{\mathcal X^{k+1}}G_0(x_0)\prod_{s=1}^{k}G_s(x_{s-1},x_s)\,Q_{0:k}(\mathrm dx_{0:k})$, $k\in[K]$. Its SMC sampler draws $N$ particles from $q$, each with weight $1$; at each $k\in[K]$ it transports them with $\mathsf M_k$, reweights them with $G_k$ and optionally resamples them.
\end{definition}

\begin{definition}[{Incremental importance weights; From \citet{arbel2021annealedflowtransportmonte}}]
\label{def:incw}
Let $\pi_k=\gamma_k/Z_k$, $k=0,\dots,K$, with $Z_0=1$. Sample $X_0\sim\pi_0$ and, for $k\ge1$, $X_k\sim\mathsf M_k(X_{k-1},\cdot)$ with a Markov kernel $\mathsf M_k$, so that $X_{0:k}$ has density $\bar\eta_k(x_{0:k})=\pi_0(x_0)\prod_{l=1}^{k}\mathsf M_l(x_{l-1},x_l)$. For backward Markov kernels $\mathsf L_l$, the extended target $\bar\pi_k(x_{0:k})=\pi_k(x_k)\prod_{l=0}^{k-1}\mathsf L_l(x_{l+1},x_l)$ has marginal $\pi_k$ at time $k$ by construction. The $\mathsf L_{k-1}$ are chosen so that the incremental importance weights
\begin{equation}
\label{eq:app-incw}
G_k(x_{k-1},x_k)=\frac{\gamma_k(x_k)\,\mathsf L_{k-1}(x_k,x_{k-1})}{\gamma_{k-1}(x_{k-1})\,\mathsf M_k(x_{k-1},x_k)}
\end{equation}
are well defined, and then $W_k(x_{0:k})\coloneqq\bar\gamma_k(x_{0:k})/\bar\eta_k(x_{0:k})=\prod_{l=1}^{k}G_l(x_{l-1},x_l)$, where $\bar\gamma_k=Z_k\bar\pi_k$ is the unnormalized joint target.
\end{definition}

\begin{proposition}[{Path measure targeted by an SMC sampler; from \citet{del2006sequential}}]
\label{prop:fk-target}
Let $\pi=\gamma/Z$ with $\gamma:\mathcal X\to\mathbb R_{>0}$, and let $(\pi_k)_{k=0}^{K}$, $\pi_k\propto\gamma_k$, be a path with $\pi_0=q$, the initial proposal, and $\pi_K=\pi$. Let $(\mathsf M_k)_{k\in[K]}$ be forward and $(\mathsf L_{k-1})_{k\in[K]}$ backward Markov kernels such that
\begin{equation}
\label{eq:app-ac}
\pi_k\otimes\mathsf L_{k-1}\ll\pi_{k-1}\otimes\mathsf M_k
\end{equation}
holds for every $k$, and set, for $k\ge1$ and with $Z_k\coloneqq\int_{\mathcal X}\gamma_k(x)\,\mathrm dx$,
\begin{equation}
\label{eq:app-kim-potential}
G_k(x_{k-1},x_k)=\frac{Z_k}{Z_{k-1}}\,\frac{\mathrm d\big(\pi_k\otimes\mathsf L_{k-1}\big)}{\mathrm d\big(\pi_{k-1}\otimes\mathsf M_k\big)}(x_{k-1},x_k),
\end{equation}
which equals \Cref{eq:app-incw} when the Radon--Nikodym derivative can be evaluated pointwise. Then the Feynman--Kac model of Definition~\ref{def:fk} with these kernels and potentials, $G_0\equiv1$ and $\gamma_0=q$, so that $Z_0=1$, targets the path measure $P_{0:K}(\mathrm dx_{0:K})=\pi(\mathrm dx_K)\prod_{k=1}^{K}\mathsf L_{k-1}(x_k,\mathrm dx_{k-1})$: the marginal of $x_K$ is $\pi$, and for $k\in[K]$ its intermediate normalizing constants (\Cref{def:fk}) are these $Z_k$.
\end{proposition}

\subsection{Incremental weights of \Cref{alg:smc}}
\label{app:theory-weight}

In \Cref{eq:weights}, the incremental weight is evaluated for the plan-indexed target sequence $\gamma_k$ of \Cref{eq:target_family}, with $U_k=-\log\gamma_k$. We use the forward and backward kernels
\begin{equation}
\label{eq:app-ula}
\mathsf M_k(\mathbf x,\mathbf x')=(4\pi\eta)^{-d/2}\exp\!\Big(-\tfrac{1}{4\eta}\big\lVert\mathbf x'-\mathbf x+\eta\nabla U_k(\mathbf x)\big\rVert^2\Big),
\qquad
\mathsf L_{k-1}(\mathbf x',\mathbf x)=\mathsf M_k(\mathbf x',\mathbf x),
\end{equation}
where $d=Hm$ is the dimension of $\mathcal U$. This is one of the kernel parametrizations considered by \citet[Remark~2]{cuin2026efficientstochasticoptimisationsequential}, who show that the forward and backward kernels can instead be indexed by $k-1$, with both choices agreeing in the continuous-time limit. For this Gaussian kernel pair, the incremental weight in \Cref{eq:app-incw} admits the closed-form expression derived in \citet[App.~A.1]{cuin2026efficientstochasticoptimisationsequential}, using
\begin{equation}
\label{eq:app-incw-planner}
\Lambda_j(\mathbf x,\mathbf x')=U_j(\mathbf x)+\tfrac12(\mathbf x'-\mathbf x)\cdot\nabla U_j(\mathbf x)+\tfrac{\eta}{4}\lVert\nabla U_j(\mathbf x)\rVert^2.
\end{equation}

A different simplification holds when the forward move is a Metropolis--Hastings transition $\mathsf K_k$ reversible with respect to $\pi_k\propto\gamma_k$. This includes MALA, using $\mathsf M_k$ as its proposal, as well as the RWM and HMC transitions defined in \Cref{app:kernel-defs}; for HMC, reversibility is understood on the extended state space. Choosing the backward kernel as the time reversal of $\mathsf K_k$,
\begin{equation}
\label{eq:app-reversal}
\mathsf L_{k-1}(\mathbf x',\mathrm d\mathbf x)=\frac{\pi_k(\mathrm d\mathbf x)\,\mathsf K_k(\mathbf x,\mathrm d\mathbf x')}{\pi_k(\mathrm d\mathbf x')},
\end{equation}
makes it equal to the forward kernel by reversibility. Substitution into \Cref{eq:app-incw} cancels the transition kernels, leaving
\begin{equation}
G_k(\mathbf x,\mathbf x')=\frac{\gamma_k(\mathbf x)}{\gamma_{k-1}(\mathbf x)}.
\end{equation}
Thus, the incremental weight depends only on the change in target density at the previous particle position, as in Eq.~(31) of \citet{del2006sequential} and \citet[Remark~2 and App.~A.2]{cuin2026efficientstochasticoptimisationsequential}. Reversibility removes the contribution of the move kernel, but the weight is generally nontrivial because the target changes with $k$.

Our particles are drawn from $q(\cdot\mid\bm{\mu}_0)$, so the initial proposal contributes the initial potential $G_0=\gamma_0/q(\cdot\mid\bm{\mu}_0)=e^{-C/\alpha}$ of \Cref{def:fk}, as in \citet[Thm.~1]{cuin2026efficientstochasticoptimisationsequential}. One may instead draw the initial particles from $\pi_0$ itself and start from unit weights, the convention of \Cref{def:incw} \citep{arbel2021annealedflowtransportmonte}. For any prescribed sequence $(\bm{\mu}_k)_{k=0}^{K}$, \Cref{prop:fk-target} therefore establishes that the marginal at step $K$ of the targeted path measure is $\pi_K\propto\gamma_{\bm{\mu}_K}$.

In \Cref{alg:smc}, however, this sequence is determined by the particle population. The scale $\alpha$ is set to the standard deviation of $C$ over the initial population and then held fixed; for the PDE cost, $\lambda_V$ in \Cref{eq:cost} is also set from the initial population, before $\alpha$ is determined. Such particle-dependent target sequences fall outside the standard fixed-target SMC guarantees: finite-$N$ guarantees require an interacting-particle analysis, as noted by \citet[Remark~3]{cuin2026efficientstochasticoptimisationsequential}. We leave this analysis open.

\section{Kernels and optimization step of \Cref{alg:smc}}
\label{app:kernel}

In this section, we derive the kernels used within the SMC step of \Cref{alg:smc}, ULA with Jarzynski weights as in JALA \citep{cuin2026learning}, and the three Metropolis--Hastings kernels that we compare with it in \Cref{app:kernel-abl}, MALA, RWM and HMC, which are the variants considered by \citet[App.~A]{cuin2026efficientstochasticoptimisationsequential}. It then gives the background for the optimization step: the two forms taken by the gradient of its objective, and their relation to EM and to MPPI.

\subsection{Kernels}
\label{app:kernel-defs}

With $\gamma_k=\gamma_{\bm{\mu}_k}$ from \Cref{eq:target_family}, the potential $U_k=-\log\gamma_k$ and its gradient are, up to an additive constant in $U_k$,
\begin{equation}
\label{eq:app-potential}
U_k(\mathbf a)=\frac{\lVert\mathbf a-\bm{\mu}_k\rVert^2}{2\sigma_p^2}+\frac{1}{\alpha}\,C(\mathbf a),
\qquad
\nabla U_k(\mathbf a)=\frac{\mathbf a-\bm{\mu}_k}{\sigma_p^2}+\frac{1}{\alpha}\,\nabla C(\mathbf a),
\end{equation}
where $\nabla C$ is obtained by differentiating through the rollout of the frozen world model. ULA and MALA both draw the proposal $\mathbf x'\sim\mathsf M_k(\mathbf x,\cdot)$ of \Cref{eq:app-ula}, which is the move of \Cref{eq:move}, and both use the ratio
\begin{equation}
\label{eq:app-mh-ratio}
r_k(\mathbf x,\mathbf x')\coloneqq\frac{\gamma_k(\mathbf x')\,\mathsf M_k(\mathbf x',\mathbf x)}{\gamma_k(\mathbf x)\,\mathsf M_k(\mathbf x,\mathbf x')}=\exp\!\big(\Lambda_k(\mathbf x,\mathbf x')-\Lambda_k(\mathbf x',\mathbf x)\big),
\end{equation}
where the second equality is the evaluation cited in \Cref{app:theory-weight}.

\paragraph{ULA with Jarzynski weights (\Cref{alg:smc}).} The proposal is always accepted, and the weight of the particle is multiplied by $r_k(\mathbf x,\mathbf x')$ (\Cref{app:theory-weight}).

\paragraph{MALA (alternative).} The proposal is accepted with probability $1\wedge r_k(\mathbf x,\mathbf x')$, and otherwise the particle stays at $\mathbf x$; this defines the transition $\mathsf K_k$ of \Cref{app:theory-weight} \citep[App.~A.2]{cuin2026efficientstochasticoptimisationsequential}. The weight is not changed by the move (\Cref{app:theory-weight}).

\paragraph{RWM (alternative).} The proposal $\mathbf x'=\mathbf x+\epsilon\,\boldsymbol\xi$, $\boldsymbol\xi\sim\mathcal N(0,I)$, does not use $\nabla U_k$ and is symmetric, so it is accepted with probability
\begin{equation}
\label{eq:app-rwm}
1\wedge\frac{\gamma_k(\mathbf x')}{\gamma_k(\mathbf x)}=1\wedge\exp\big(U_k(\mathbf x)-U_k(\mathbf x')\big),
\end{equation}
and otherwise the particle stays at $\mathbf x$. This proposal--acceptance rule defines a transition reversible with respect to $\pi_k$ \citep[App.~A.2]{cuin2026efficientstochasticoptimisationsequential}, and the weight is not changed by the move (\Cref{app:theory-weight}).

\paragraph{HMC (alternative).} Following \citet[App.~A.2]{cuin2026efficientstochasticoptimisationsequential}, the transition acts on the extended state $(\mathbf x,\mathbf p)\in\mathcal U\times\mathcal U$, with extended unnormalized target $\widetilde\gamma_k(\mathbf x,\mathbf p)=\gamma_k(\mathbf x)\exp(-\tfrac12\mathbf p^{\top}M^{-1}\mathbf p)$ and Hamiltonian $\mathcal H_k(\mathbf x,\mathbf p)=U_k(\mathbf x)+\tfrac12\mathbf p^{\top}M^{-1}\mathbf p$ for a positive-definite mass matrix $M$, which they take to be a multiple of the identity. The momentum is refreshed at the start of the move, $\mathbf p_0\sim\mathcal N(0,M)$, and $L$ leapfrog steps of size $\epsilon$ from $(\mathbf x_0,\mathbf p_0)=(\mathbf x,\mathbf p_0)$,
\begin{equation}
\label{eq:app-leapfrog}
\mathbf p_{l+\frac12}=\mathbf p_l-\tfrac{\epsilon}{2}\nabla U_k(\mathbf x_l),\qquad
\mathbf x_{l+1}=\mathbf x_l+\epsilon\,M^{-1}\mathbf p_{l+\frac12},\qquad
\mathbf p_{l+1}=\mathbf p_{l+\frac12}-\tfrac{\epsilon}{2}\nabla U_k(\mathbf x_{l+1}),
\end{equation}
give the proposal $(\mathbf x_L,\mathbf p_L)$. The next extended state is $(\mathbf x_L,-\mathbf p_L)$ with probability $1\wedge\exp\big(\mathcal H_k(\mathbf x,\mathbf p_0)-\mathcal H_k(\mathbf x_L,\mathbf p_L)\big)$ and $(\mathbf x,\mathbf p_0)$ otherwise. The momentum is refreshed at the start of each move, so the weight is not changed by the move (\Cref{app:theory-weight}).
Each HMC move costs $L$ evaluations of $C$ and $\nabla C$, against one for each of the other three kernels.

\subsection{Background}
\label{app:em}

This section gives the context for the optimization step of \Cref{eq:mstep} and for the second estimator of its gradient, \Cref{eq:mstep-path}. With $\gamma_{\bm{\mu}}$ and $\pi_{\bm{\mu}}$ the target of \Cref{eq:target_family}, let
\begin{equation}
\label{eq:app-ell-beta}
\ell(\bm{\mu})\coloneqq-\log\int_{\mathcal U}\gamma_{\bm{\mu}}(\mathbf a)\,\mathrm d\mathbf a
=-\log\E_{q(\cdot\mid\bm{\mu})}\big[e^{-C(\mathbf a)/\alpha}\big],
\end{equation}
the objective of \Cref{sec:latent_smc_planning}. We assume that $C$ is bounded below, so that this integral is finite and positive.

\paragraph{The two gradients.} For the Gaussian reference of \Cref{eq:target_family}, the potential $U_{\bm{\mu}}=-\log\gamma_{\bm{\mu}}$ has $\nabla_{\bm{\mu}}U_{\bm{\mu}}(\mathbf a)=-(\mathbf a-\bm{\mu})/\sigma_p^{2}$, so Fisher's identity, cited in \Cref{sec:latent_smc_planning}, reads
\begin{equation}
\label{eq:app-em-score}
\nabla\ell(\bm{\mu})=\E_{\pi_{\bm{\mu}}}\big[\nabla_{\bm{\mu}}U_{\bm{\mu}}(\mathbf a)\big]
=-\frac{1}{\sigma_p^{2}}\big(\E_{\pi_{\bm{\mu}}}[\mathbf a]-\bm{\mu}\big),
\end{equation}
the expectation that $\mathbf g_k$ of \Cref{eq:mstep} estimates. Writing instead $\mathbf a=\bm{\mu}+\sigma_p\bm{\epsilon}$ with $\bm{\epsilon}\sim\mathcal N(0,I)$ and differentiating under the expectation, which needs $C$ differentiable and the interchange to be valid (the \emph{pathwise assumption}), gives a second form of the same gradient and a second estimator,
\begin{equation}
\label{eq:mstep-path}
\nabla\ell(\bm{\mu})=\frac{1}{\alpha}\,\E_{\pi_{\bm{\mu}}}\big[\nabla C(\mathbf a)\big],
\qquad
\tilde{\mathbf g}_k=\textstyle\frac{1}{\alpha}\sum_{i=1}^{N}w_k^{(i)}\,\nabla C\big(\mathbf a_k^{(i)}\big),
\end{equation}
which reuses the cost gradients that the move of \Cref{eq:move} computes in any case. Both $\mathbf g_k$ and $\tilde{\mathbf g}_k$ are weighted averages over the population. \Cref{app:abl-mstep} compares them in the settings of \Cref{tab:planners}.

\paragraph{Relation to EM and to MPPI.} By \Cref{eq:app-em-score}, a step of size $\sigma_p^{2}$ along the exact gradient sets $\bm{\mu}_{k+1}=\E_{\pi_{\bm{\mu}_k}}[\mathbf a]$, which minimizes $\KL\big(\pi_{\bm{\mu}_k}\,\|\,q(\cdot\mid\bm{\mu})\big)$ over $\bm{\mu}$ and is therefore the M-step of EM for this family, as well as the minimization of \citet[Eq.~(9)]{williams2016aggressive} with $\pi_{\bm{\mu}_k}$ in place of their optimal measure. With $N$ particles the same step gives $\bm{\mu}_{k+1}=\sum_{i=1}^{N}w_k^{(i)}\mathbf a_k^{(i)}$, and for particles drawn afresh from $q(\cdot\mid\bm{\mu}_k)$ and weighted by $e^{-C/\alpha}$ this is the MPPI update \citep[Eqs.~(33)--(34)]{williams2016aggressive}. Unlike their optimal measure, $\pi_{\bm{\mu}_k}$ depends on the plan, so each update of the plan changes the target of the next SMC step.

\section{Implementation details}
\label{app:impl}

The following table provides implementation details for every planner in  \Cref{tab:planners}.

\begin{table}[!htbp]
  \caption{Settings of our planner. Where a value differs between environments, it is listed per environment.}
  \label{tab:hyper}
  \centering
  \setlength{\tabcolsep}{5pt}
  \small
  \begin{tabular}{llll}
    \toprule
    Group & Parameter & Value & Note \\
    \midrule
    \multirow{3}{*}{Kernel}
      & move                        & ULA, Jarzynski weights & \Cref{eq:move,eq:weights} \\
      & step size $\eta$            & $0.18$     & fixed \\
      & reference width $\sigma_p$  & $1$        & \Cref{eq:target_family} \\
    \midrule
    \multirow{4}{*}{SMC}
      & particles $N$               & $12$       & all environments \\
      & ESS fraction $\tau$         & $0.5$      & resample threshold $\tau N$ \\
      & tempering $\beta_k$         & adaptive   & \Cref{app:tempering} \\
      & temperature $\alpha$        & $\operatorname{std}C$ & initial population \\
    \midrule
    M-step
      & optimizer \textsc{opt}     & Adam       & learning rate $0.1$ \\
    \midrule
    \multirow{2}{*}{Cost}
      & action regularizer $\lambda_a$ & $0.1$   & \Cref{eq:cost} \\
      & PDE weight $\lambda_V$      & $\operatorname{std}\mathcal L/\operatorname{std}C_V$ & initial population \\
    \midrule
    \multirow{7}{*}{Roadmap}
      & solver                      & heat       & \Cref{eq:heat} \\
      & diffusion constant $c$      & $1$ / $30$ / $10$ & Point maze / Wall / Push-T \\
      &                             & $3$ / $1$  & Reacher / Cube \\
      & $\kappa$-NN                 & $4$ / $32$ / $64$ & Point maze / Wall / Push-T \\
      &                             & $16$       & Reacher, Cube \\
      & graph nodes                 & $3.8$k / $12.6$k & Point maze / Wall \\
      &                             & $55$k      & Push-T, Reacher, Cube \\
    \midrule
    Budget & iterations $K$         & $150$      & all planners of \Cref{tab:planners} \\
    \bottomrule
  \end{tabular}
\end{table}

\paragraph{Roadmap and cost-to-go.} The roadmap of \Cref{def} is built once per environment and shared across goals and seeds. Its vertices are drawn from a buffer of encoded transitions. On Reacher, Push-T and Cube the buffer comes from the training split of the dataset, which excludes the held-out trajectories that give the starts and goals of Push-T and Cube; on Point maze and Wall, whose starts and goals are drawn at random, it comes from $200$ of the $2000$ episodes of the dataset and from a random subset of $1400$ of its $1920$ episodes. The vertices are all of the buffer on Point maze ($3.8$k transitions) and Wall ($12.6$k), and a fixed uniform subsample of $55$k on Push-T, Reacher and Cube. Coordinates are standardized with the mean and variance of all encoded states and successors. The $\kappa$-nearest-neighbor search is exact on Point maze; on the larger roadmaps it merges exact searches over random candidate subsets of $4096$ vertices, keeping each vertex at most once among the neighbors of a successor. Bandwidths are clamped below at $10^{-3}$. \Cref{eq:heat} is solved by Jacobi-preconditioned conjugate gradient started from zero, with relative tolerance $10^{-8}$ and at most $5000$ iterations; entries of the returned iterate that are not positive are set to zero before \Cref{eq:rank}, where they share one averaged rank.

\paragraph{Calibration of the roadmap.} For the diffusion time, \citet{crane2012geodesicsheat} choose $t=ch^2$, where $h$ is the mean spacing between adjacent nodes, for a Laplacian that carries a $1/h^2$ normalization. Our $L$ carries no such normalization, and its diagonal is the degree $D_{ii}$. We therefore set $t=c/\operatorname{median}_iD_{ii}$ (\Cref{sec:cost}), so that $tL$ is dimensionless and of order $c$ at a typical vertex, and choose $\kappa$ and the diffusion constant $c$ per environment from the offline dataset alone. The dataset stores whole episodes, so for a goal vertex $g$ and an earlier vertex $i$ of the same episode, the number of steps from $i$ to $g$ is an observed cost-to-go. We score each $(\kappa,c)$ on a grid by the Spearman rank correlation between $V_i$ and this number of steps over the earlier vertices of the episode, averaged over $60$ goal vertices. The best settings differ by less than one standard error over goals, so we treat all settings within one standard error of the grid maximum as tied, prefer among them settings at which the solution of \Cref{eq:heat} is positive at every vertex, and check the chosen setting on all states of the episodes of $20$ goal vertices drawn at random from the roadmap, scored in the same way; on Cube, two tied settings were compared in this way and the one with the higher correlation kept; no planning seed, world-model rollout or success signal enters this choice. The chosen values are listed in \Cref{tab:hyper}.

\paragraph{Tempering.}\label{app:tempering}
Annealed importance sampling \citep{neal1998annealedimportancesampling} and SMC samplers \citep{del2006sequential} approach a tilted target through tempered intermediates rather than in one step, and \citet{arbel2021annealedflowtransportmonte} set the temperature of each intermediate from the ESS of the particles, a schedule studied further by \citet{syed2025optimisedannealedsequentialmonte} and applied to trajectory optimization by \citet{yang2026tempered}. We adopt that device unchanged, as a warm-up on the target family, denoted by 
\begin{equation}
\label{eq:tempered}
\gamma_{\bm{\mu},\beta}(\mathbf a)=q(\mathbf a\mid\bm{\mu})\exp\!\big(-\beta\,C(\mathbf a)/\alpha\big),
\qquad \pi_{\bm{\mu},\beta}\propto\gamma_{\bm{\mu},\beta},\qquad 0\le\beta\le1,
\end{equation}
Under this setting, we can replace $\gamma_k$ by $\gamma_{\bm{\mu}_k,\beta_k}$, start at $\beta_0=0$, where the reference is the first target, so that the initial potential of \Cref{alg:smc} is replaced by $W_0^{(i)}=1$, and raise $\beta_k$ by bisection on $[\beta_{k-1},1]$ to the value at which the ESS after the change of target equals $N/2$, taking $\beta_k=1$ when that ESS stays above $N/2$ and $\beta_k=\beta_{k-1}$ when it is already below. The change of target of \Cref{eq:weights} then carries the further factor $\exp(-(\beta_k-\beta_{k-1})C(\mathbf x)/\alpha)$ alongside the Gaussian ratio, for which the cost at $\mathbf x$ is already available. 

\paragraph{Weights.} We store the particle weights in log space, adding each factor through its logarithm. Normalized weights, ESS, and resampling probabilities are obtained by applying softmax to the log weights. We subtract the maximum log weight before exponentiation to avoid overflow and ensure a nonzero normalization sum. We set $\alpha$ to the standard deviation of $C$ over the initial population, thereby scaling the initial cost variation to unit standard deviation across environments.

\paragraph{Wall-clock cost of the PDE cost.}
\Cref{tab:pde-time} reports the overhead of the PDE cost relative to the same planner using the $L_2$ cost ($\lambda_V=0$, with no roadmap loaded or built). We measure one seed per environment at $H{=}10$ ($H{=}5$ on Cube), using 30 iterations and the pathwise M-step of \Cref{eq:mstep-path} in every environment, including Point maze. Measurements are taken on an NVIDIA RTX 5080 shared with another process; the cost of a 150-iteration trial is extrapolated from the per-iteration time. Since both configurations use the same planner, the caveat in \Cref{sec:experiments} about comparing wall-clock times across implementations does not apply.

The roadmap is built once per environment, in at most $3$\,s. For each goal, the PDE cost requires solving \Cref{eq:heat} (at most $1$\,s) and fitting $V$ (2--13\,s), which accounts for most of the overhead. Evaluating the initial population to set $\lambda_V$ (\Cref{sec:cost}) takes under $0.4$\,s. Each iteration adds a forward and backward pass of $V$ over the particles' predicted latents; the resulting per-iteration times differ by less than the variation between runs.

Relative to a 150-iteration trial with the $L_2$ cost, the per-goal overhead is $1.4\%$ on Point maze, $6$--$7\%$ on Wall and Push-T, and $35\%$ and $65\%$ on Reacher and Cube, respectively. The lower overhead on Point maze, Wall, and Push-T reflects the longer DINO-WM iteration time ($0.34$--$1.3$\,s), compared with LeWM ($0.14$--$0.23$\,s) on Reacher and Cube.

\begin{table}[!htbp]
  \caption{Wall-clock time, one seed per environment: roadmap (per environment), solve and fit (per goal), iteration with each cost, per-goal overhead over a $150$-iteration $L_2$ trial.}
  \label{tab:pde-time}
  \centering
  \small
  \setlength{\tabcolsep}{5pt}
  \begin{tabular}{lrrrrrr}
    \toprule
     & Roadmap (s) & Solve (s) & Fit (s) & Iter.\ $L_2$ (ms) & Iter.\ PDE (ms) & Overhead \\
    \midrule
    Point maze & $0.31$ & $<0.01$ & $2.34$ & $1335$ & $1305$ & $1.4\%$ \\
    Wall       & $0.42$ & $0.59$ & $2.67$ & $337$ & $359$ & $7.1\%$ \\
    Push-T     & $1.39$ & $0.99$ & $10.95$ & $1300$ & $1334$ & $6.2\%$ \\
    Reacher    & $2.84$ & $0.41$ & $11.39$ & $231$ & $198$ & $35\%$ \\
    Cube       & $2.05$ & $0.17$ & $13.09$ & $138$ & $138$ & $65\%$ \\
    \bottomrule
  \end{tabular}
\end{table}

\section{Choice of the operator on the roadmap}
\label{app:operator}
\paragraph{Related graph PDEs.}
We consider other graph-based cost structures related to the heat equation. Shortest-path distance on a roadmap \citep{kavraki1996probabilistic} is the graph counterpart of the Eikonal cost-to-go \citep{kimmel2001optimal,mirebeau2019hamiltonian}. Graphs constructed from offline buffers have also been used to search for waypoints \citep{eysenbach2019search,emmons2020sparse}; here, we use the roadmap to define the cost itself. Another option is to use $p$-Laplacian extensions of labelled values, which interpolate between harmonic ($p=2$) and absolutely minimal Lipschitz ($p\to\infty$) extensions \citep{nadler2009semi,calder2019game,zhang2026planner}. With only the goal vertex labelled, these extensions are constant. To obtain nontrivial costs, we assign a unit source to every other vertex. The resulting solutions lie between the two limits of \Cref{eq:heat}, but require a convex minimization for each goal.

\paragraph{The heat step.}
\Cref{eq:heat} computes one backward Euler step of the heat equation from a point source \citep[Eq.~(3)]{crane2012geodesicsheat}. On a connected roadmap, the resulting field $u$ strictly decreases away from the goal along at least one outgoing edge. Thus, $u$ is a reachability field rather than a cost, and the induced potential $V$ has no local minimum on the roadmap other than the goal. We use only the ordering induced by $u$: converting its magnitude to distance through the Varadhan estimate $\sqrt{-4t\log u}$ is sensitive to errors in the field values \citep{crane2012geodesicsheat}, whereas the ordering is invariant under monotone transformations.

\paragraph{Offline comparison.}
\Cref{tab:calib-operators} compares \Cref{eq:heat} with four alternative orderings on the same roadmap: the number of edges to the goal, the Green's function, shortest-path distance and Euclidean distance $\lVert\xi_i-\xi_g\rVert$, using the criterion from the selection of $\kappa$ and $c$ (\Cref{app:impl}) on 60 additional goals. Among goals for which both orderings are defined, \Cref{eq:heat} outperforms both limits in all five environments, by $0.008$--$0.035$ over the number of edges and $0.007$--$0.037$ over the Green's function; only the former difference on Cube exceeds two standard errors. Its difference relative to shortest-path distance changes sign across environments, from $-0.012\pm0.017$ on Wall to $+0.062\pm0.017$ on Point maze. Differences are generally larger on the goals used to select $\kappa$ and $c$, and some change sign. Thus, the offline data do not establish a consistent advantage for \Cref{eq:heat} over the graph-based alternatives.

All roadmap orderings outperform Euclidean distance by at least $0.07$ on Reacher, Point maze, and Cube. On Wall, the advantage of \Cref{eq:heat} is $0.051\pm0.027$, while on Push-T all five orderings lie within $0.05$ of one another. The largest advantages over Euclidean distance occur on Point maze and Reacher, the same environments where the PDE cost improves on the $L_2$ cost (\Cref{sec:exp:ablation}). This improvement is statistically significant after Holm correction only on Reacher; on Point maze, it is significant only with the score form of the optimization step (\Cref{app:abl-mstep}). With only five environments, this correspondence is suggestive rather than conclusive.

\begin{table}[!htbp]
  \caption{Spearman correlation (mean $\pm$ standard error over $60$ held-out goals; criterion of \Cref{app:impl}) of each ordering of the roadmap vertices with the observed number of steps to the goal (top), and heat step$^{\ast}$ (default) minus each ordering (bottom). Shortest path: Dijkstra distance with edge lengths $\lVert\xi_i-\xi_j\rVert$. Number of edges omits goals where it is constant (Reacher, Wall).}
  \label{tab:calib-operators}
  \centering
  \scriptsize
  \setlength{\tabcolsep}{3.5pt}
  \begin{tabular}{lccccc}
    \toprule
    Ordering & Reacher & Point maze & Wall & Push-T & Cube \\
    \midrule
    \multicolumn{6}{l}{\emph{Rank correlation with the number of steps to the goal}} \\
    Heat step$^{\ast}$ & $0.545\pm0.060$ & $0.807\pm0.033$ & $0.824\pm0.036$ & $0.595\pm0.062$ & $0.832\pm0.022$ \\
    Number of edges & $0.552\pm0.061$ & $0.794\pm0.031$ & $0.800\pm0.029$ & $0.568\pm0.063$ & $0.819\pm0.022$ \\
    Green's function & $0.538\pm0.060$ & $0.780\pm0.041$ & $0.787\pm0.040$ & $0.583\pm0.062$ & $0.822\pm0.020$ \\
    Shortest path & $0.491\pm0.066$ & $0.745\pm0.038$ & $0.835\pm0.037$ & $0.601\pm0.060$ & $0.826\pm0.022$ \\
    Euclidean distance & $0.337\pm0.063$ & $0.422\pm0.058$ & $0.772\pm0.049$ & $0.608\pm0.058$ & $0.747\pm0.031$ \\
    \midrule
    \multicolumn{6}{l}{\emph{Heat step minus operator, paired over goals}} \\
    Number of edges & $+0.008\pm0.023$ & $+0.013\pm0.007$ & $+0.035\pm0.020$ & $+0.027\pm0.025$ & $+0.014\pm0.006$ \\
    Green's function & $+0.007\pm0.023$ & $+0.027\pm0.017$ & $+0.037\pm0.029$ & $+0.012\pm0.033$ & $+0.010\pm0.012$ \\
    Shortest path & $+0.054\pm0.037$ & $+0.062\pm0.017$ & $-0.012\pm0.017$ & $-0.006\pm0.036$ & $+0.006\pm0.010$ \\
    Euclidean distance & $+0.208\pm0.066$ & $+0.385\pm0.051$ & $+0.051\pm0.027$ & $-0.013\pm0.037$ & $+0.085\pm0.025$ \\
    \bottomrule
  \end{tabular}
\end{table}

\section{Scope of the PDE cost}
\label{app:pde-scope}
$V$ is informative only where the offline transitions cover the latent space that a plan traverses, and only about the part of the state that the roadmap latent encodes. A missing passage, an edge that the dynamics do not realize, or a latent that places distant states close together can make $V$ misorder states. $V$ is also evaluated at predicted latents, which may lie away from every vertex, where the roadmap does not constrain it (\Cref{app:operator}). Nothing in \Cref{alg:smc} tests $V$ during planning; $C_V$ enters $C$ with the weight $\lambda_V$ set once from the initial population (\Cref{sec:cost}). Each roadmap of \Cref{tab:hyper} is connected, which does not show whether it misses or adds a passage. On $12$ of the $50$ Cube goals, the solver of \Cref{app:impl} returns an iterate that is not positive at up to about $5\%$ of the vertices, which then share the largest $V$; our planner solves the same $10$ of these $12$ seeds with either cost.

\section{Baseline configurations}
\label{app:baselines}

\paragraph{GD.} GD optimizes the action sequence by backpropagating through the world model \citep[Sec.~4.1]{psenka2026parallelstochasticgradientbasedplanning}. We run Adam with step size $0.1$ on the action sequence, initialized from $\mathcal{N}(0,1)$, on the goal distance (\Cref{sec:cost}), through the full $H$-step rollout. On Cube, the sequence is initialized at zero and the goal distance is augmented by the action regularizer of \Cref{eq:cost} with weight $0.1$, decreasing linearly to zero. The released implementation optimizes a single particle (\Cref{app:abl-baselines}). In \Cref{tab:planners}, GD rolls out $12$ candidate sequences per iteration, and the Adam step averages the gradient of the goal distance over them. The executed plan is the optimized sequence.

\paragraph{GRASP.} We use the published configuration of \citet{psenka2026parallelstochasticgradientbasedplanning}: joint optimization over actions and intermediate latent states with a quadratic penalty, state step size $0.1$, action step size $10^{-3}$, and periodic full-gradient synchronization (\texttt{gd\_interval}$=100$, \texttt{gd\_opt\_steps}$=25$). With $12$ particles, the released implementation optimizes $12$ independently initialized and perturbed copies of this problem. At a synchronization, it refines the copy of lowest terminal dynamics loss, the squared distance between the world model's prediction from the last intermediate state and the goal, by full-gradient steps on the goal distance of the rollout from the start. If it decreases, the refined actions and their rollout states replace all $12$ copies; otherwise the refinement step size is halved. With $150$ iterations, this happens once, after iteration $100$. Where the released implementation executes the first copy, we execute after every iteration the copy of lowest terminal dynamics loss, the criterion of its synchronization, which reuses that iteration's forward pass.

\paragraph{CEM.} CEM \citep{zhou2025dinowmworldmodelspretrained, maes2026leworldmodelstableendtoendjointembedding} draws $12$ samples and re-fits to $3$ elites at each of the $150$ iterations. In \Cref{app:abl-baselines}, it draws $30$ samples with $3$ elites for $5$ iterations, the $150$ world-model rollouts per trial that a single-particle planner performs in $150$ iterations.

\paragraph{MPPI.} We run MPPI \citep{williams2016aggressive} on the goal distance with the unchanged solver of the planning library released with LeWM \citep{maes2026leworldmodelstableendtoendjointembedding}. Each iteration draws $12$ sequences, the current mean and perturbations of it by independent standard normal noise of fixed scale $1$, and replaces the mean by their average weighted by a softmax of the negative goal distance at temperature $0.5$; all $12$ are weighted, since the solver averages over its $30$ lowest-cost samples. The mean starts at zero and is the executed plan. The solver's defaults are $300$ samples and $30$ iterations.

\paragraph{LeFlow.} LeFlow \cite{huang2026leflow} generates a latent path with a flow model and decodes it into actions with an inverse dynamics model. GC-IDM \cite{nguyen2026latent} replaces search by a goal-conditioned inverse dynamics policy. LEAP \cite{pham2026latent} refines a learned proposal through the world model. Hierarchical world models \cite{zhang2026hierarchical} plan over subgoals with a high-level predictor. We run LeFlow and GC-IDM with their released code, which uses LeWM, on its two environments, Reacher and Cube (\Cref{tab:learned}).

\section{Ablations}
\label{app:ablations}

\subsection{Propagation kernel}
\label{app:kernel-abl}

\begin{figure}[!htbp]
  \centering
  \includegraphics[width=\linewidth]{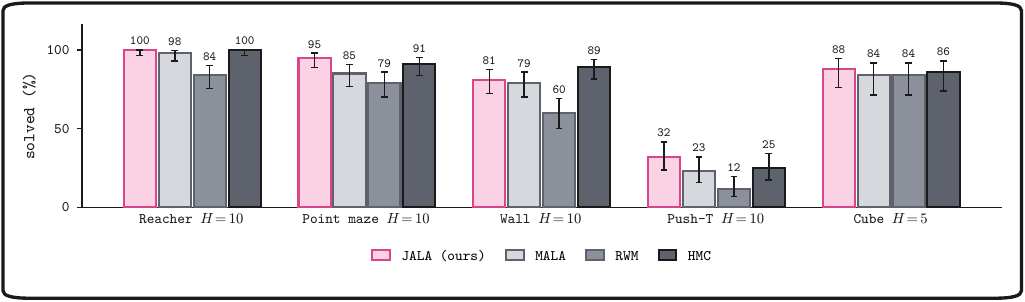}
  \caption{Success rate by propagation kernel, with every other component of the planner fixed: $100$ paired seeds at $H{=}10$ on Reacher, Point maze, Wall and Push-T, and $50$ seeds at $H{=}5$ on Cube. JALA is the kernel of \Cref{alg:smc}; its bars repeat \Cref{tab:planners} on Reacher, Wall and Push-T and \Cref{sec:exp:planning} on Cube. On Point maze, every kernel uses the pathwise optimization step of \Cref{eq:mstep-path}, so the JALA bar ($95\%$) is not the score-form cell of \Cref{tab:planners} ($98\%$). Error bars are $95\%$ Wilson intervals for each rate separately; significance is the paired McNemar test of \Cref{app:ci}.}
  \label{fig:kernels}
\end{figure}

\Cref{fig:kernels} replaces the ULA move of \Cref{alg:smc} by each Metropolis--Hastings kernel of \Cref{app:kernel-defs}, the variants of \citet[App.~A]{cuin2026efficientstochasticoptimisationsequential}. All other settings are those of \Cref{tab:hyper}, except that the optimization step takes the pathwise form of \Cref{eq:mstep-path} in every environment, Point maze included. We use the seeds of \Cref{tab:planners} at $H{=}10$ on Reacher, Point maze, Wall and Push-T, and the $50$ seeds of Cube.

\paragraph{Step sizes and mass matrix.} ULA keeps the fixed $\eta=0.18$ of \Cref{tab:hyper}, as acceptance-rate tuning does not apply to unadjusted kernels \citep{kim2025tuningsequentialmontecarlo}, so only the Metropolis--Hastings kernels receive a step-size adaptation. For MALA, RWM and HMC, the step size starts at $0.18$ and is adapted at every iteration by dual averaging \citep{hoffman2014no}, toward a population-averaged acceptance probability of $0.574$, $0.234$ and $0.651$, respectively. Dual averaging shrinks its iterates toward ten times the initial step size, so these kernels accept few proposals at first. Only HMC is preconditioned, with a diagonal mass matrix $\propto\lambda+\bar v^{1/2}$ of unit geometric mean, where $\lambda=0.05$ and $\bar v$ is an exponential moving average over iterations (factor $0.7$) of the population-averaged squared cost gradient \citep{li2016preconditioned}. Its number of leapfrog steps is drawn uniformly from $\{1,\dots,10\}$ at each iteration, so an HMC move costs $5.5$ evaluations of $C$ and $\nabla C$ on average, against one for the other kernels.

\paragraph{Results.} JALA has the highest success rate in four of the five settings, tied with HMC on Reacher. On Wall, HMC reaches $89\%$ against $81\%$, not significantly ($12$ seeds against $4$, exact McNemar test, $p=0.077$; \Cref{app:ci}). MALA differs from JALA by its accept--reject step (\Cref{app:kernel-defs}) and its adapted step size. JALA succeeds at least as often as MALA in every setting, significantly on Point maze ($10$ seeds against $0$, $p=0.002$) and nominally on Push-T ($13$ against $4$, $p=0.049$, not significant after a Holm correction). RWM, the only kernel that does not use $\nabla U_k$, has the lowest success rate in every setting, tied with MALA on Cube, and is significantly behind JALA on Reacher, Point maze, Wall and Push-T ($p\le0.001$). HMC trails JALA by $2$ to $7$ points on Point maze, Push-T and Cube, none significantly, at $5.5$ times the cost per move.

\subsection{Iteration budget}
\label{app:abl-budget}

No setting of our planner depends on the number of iterations $K$, so the runs of \Cref{tab:planners}, which stop at the first success, give its success rate at every budget up to $K=150$ (\Cref{tab:abl-budget}; per iteration in \Cref{fig:success,fig:curves_rest}). With $K=50$, our planner attains at least $80\%$ of its success at $K=150$ in ten of the thirteen settings, all but Reacher at $H{=}15$ and Push-T at $H{=}10$ and $H{=}15$.

\begin{table}[!htbp]
  \caption{Success rate (\%) of our planner within $K$ iterations, from the runs of \Cref{tab:planners} and, for Cube, \Cref{sec:exp:planning}.}
  \label{tab:abl-budget}
  \centering
  \small
  \begin{tabular}{llrrrrr}
    \toprule
    Environment & $H$ & $K=10$ & $K=25$ & $K=50$ & $K=100$ & $K=150$ \\
    \midrule
    Reacher & $5$ & $67$ & $94$ & $99$ & $100$ & $100$ \\
    Reacher & $10$ & $33$ & $67$ & $84$ & $99$ & $100$ \\
    Reacher & $15$ & $23$ & $49$ & $67$ & $85$ & $89$ \\
    Point maze & $5$ & $60$ & $70$ & $80$ & $86$ & $88$ \\
    Point maze & $10$ & $82$ & $88$ & $95$ & $97$ & $98$ \\
    Point maze & $15$ & $77$ & $89$ & $94$ & $95$ & $96$ \\
    Wall & $5$ & $38$ & $71$ & $78$ & $89$ & $92$ \\
    Wall & $10$ & $54$ & $64$ & $69$ & $79$ & $81$ \\
    Wall & $15$ & $55$ & $68$ & $74$ & $80$ & $84$ \\
    Push-T & $5$ & $41$ & $68$ & $86$ & $92$ & $93$ \\
    Push-T & $10$ & $7$ & $12$ & $16$ & $26$ & $32$ \\
    Push-T & $15$ & $0$ & $1$ & $2$ & $2$ & $4$ \\
    Cube & $5$ & $82$ & $88$ & $88$ & $88$ & $88$ \\
    \bottomrule
  \end{tabular}
\end{table}

\subsection{Baselines with other particle counts}
\label{app:abl-baselines}

\Cref{tab:abl-1particle} runs GD and GRASP with a single particle for $150$ iterations. \Cref{tab:abl-cem} runs CEM with $30$ samples and $3$ elites for $5$ iterations, $150$ world-model rollouts, as many as a single particle performs in $150$ iterations. Averaged over the twelve cells, GD succeeds in $36.7\%$ of trials with a single particle and in $50.9\%$ with $12$ candidates (\Cref{tab:planners}), and GRASP in $26.9\%$ and $29.9\%$. CEM succeeds in $40.7\%$ with $30$ samples for $5$ iterations and in $45.3\%$ with $12$ samples for $150$ iterations.

\begin{table}[!htbp]
  \caption{Success rate (\%) and mean iterations to success, as in \Cref{tab:planners}, for GD and GRASP with a single particle; ours repeats \Cref{tab:planners}. Bold: best per column.}
  \label{tab:abl-1particle}
  \centering
  \setlength{\tabcolsep}{7pt}
  \renewcommand{\arraystretch}{1.15}
  \resizebox{1\textwidth}{!}{%
  \begin{tabular}{lcccccccccccc}
    \toprule
    & \multicolumn{3}{c}{Reacher} & \multicolumn{3}{c}{Point maze} & \multicolumn{3}{c}{Wall} & \multicolumn{3}{c}{Push-T} \\
    \cmidrule(lr){2-4} \cmidrule(lr){5-7} \cmidrule(lr){8-10} \cmidrule(lr){11-13}
    Planner & 5 & 10 & 15 & 5 & 10 & 15 & 5 & 10 & 15 & 5 & 10 & 15 \\
    \midrule
    GD & 57 \tiny{(11)} & 38 \tiny{(15)} & 16 \tiny{(19)} & 54 \tiny{(23)} & 51 \tiny{(13)} & 50 \tiny{(23)} & 28 \tiny{(30)} & 41 \tiny{(35)} & 25 \tiny{(22)} & 62 \tiny{(23)} & 16 \tiny{(33)} & 2 \tiny{(26)} \\
    GRASP & 27 \tiny{(91)} & 33 \tiny{(57)} & 32 \tiny{(52)} & 44 \tiny{(63)} & 49 \tiny{(37)} & 62 \tiny{(34)} & 21 \tiny{(103)} & 9 \tiny{(110)} & 5 \tiny{(108)} & 37 \tiny{(106)} & 4 \tiny{(101)} & 0 \\
    \midrule
    \textbf{Ours} & \best{100} \tiny{(10)} & \best{100} \tiny{(25)} & \best{89} \tiny{(33)} & \best{88} \tiny{(16)} & \best{98} \tiny{(9)} & \best{96} \tiny{(9)} & \best{92} \tiny{(23)} & \best{81} \tiny{(19)} & \best{84} \tiny{(19)} & \best{93} \tiny{(20)} & \best{32} \tiny{(53)} & \best{4} \tiny{(73)} \\
    \bottomrule
  \end{tabular}}
\end{table}

\begin{table}[!htbp]
  \caption{Success rate (\%) and mean iterations to success, as in \Cref{tab:planners}, for CEM with $30$ samples and $3$ elites for $5$ iterations; ours repeats \Cref{tab:planners}. Bold: best.}
  \label{tab:abl-cem}
  \centering
  \setlength{\tabcolsep}{7pt}
  \renewcommand{\arraystretch}{1.15}
  \resizebox{1\textwidth}{!}{%
  \begin{tabular}{lcccccccccccc}
    \toprule
    & \multicolumn{3}{c}{Reacher} & \multicolumn{3}{c}{Point maze} & \multicolumn{3}{c}{Wall} & \multicolumn{3}{c}{Push-T} \\
    \cmidrule(lr){2-4} \cmidrule(lr){5-7} \cmidrule(lr){8-10} \cmidrule(lr){11-13}
    Planner & 5 & 10 & 15 & 5 & 10 & 15 & 5 & 10 & 15 & 5 & 10 & 15 \\
    \midrule
    CEM & 62 \tiny{(3)} & 38 \tiny{(2)} & 28 \tiny{(3)} & 51 \tiny{(2)} & 67 \tiny{(2)} & 69 \tiny{(3)} & 33 \tiny{(4)} & 40 \tiny{(4)} & 49 \tiny{(4)} & 45 \tiny{(4)} & 4 \tiny{(5)} & 2 \tiny{(4)} \\
    \midrule
    \textbf{Ours} & \best{100} \tiny{(10)} & \best{100} \tiny{(25)} & \best{89} \tiny{(33)} & \best{88} \tiny{(16)} & \best{98} \tiny{(9)} & \best{96} \tiny{(9)} & \best{92} \tiny{(23)} & \best{81} \tiny{(19)} & \best{84} \tiny{(19)} & \best{93} \tiny{(20)} & \best{32} \tiny{(53)} & \best{4} \tiny{(73)} \\
    \bottomrule
  \end{tabular}}
\end{table}

\subsection{Effect of the Cost Structure on Other Planners}
\label{app:abl-cost}
\begin{figure}[h!]
    \centering
    \includegraphics[width=0.5\linewidth]{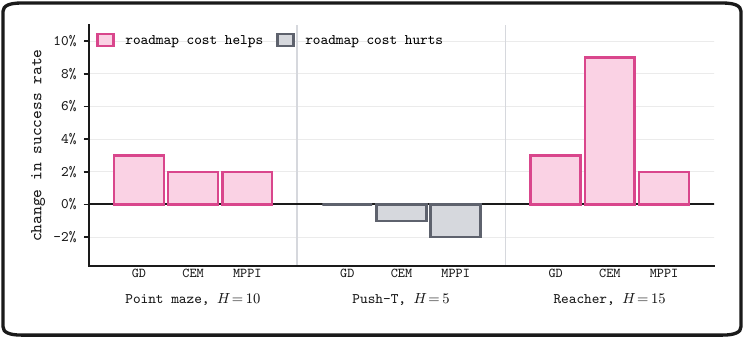}
    \caption{Performance gains from the PDE-based cost over the $L_2$ cost across planners and environments.}
    \label{fig:pde_baseline_gain}
\end{figure}

In \cref{fig:pde_baseline_gain}, we examine whether the PDE-based cost also benefits planners beyond SMC. Its effects are substantially smaller for these planners, although the trends are broadly consistent: performance tends to improve on Point maze and Reacher, but decreases slightly on Push-T. Overall, the changes are marginal, indicating that the benefit of the PDE cost is more pronounced within our SMC-based planning procedure. Notably, SMC also achieves higher average success than the other planners when using the $L_2$ cost, suggesting that the observed gains reflect both the underlying planner and its interaction with the cost structure.

\paragraph{Form of the optimization step.}\label{app:abl-mstep} \Cref{tab:abl-mstep} compares the pathwise form of \Cref{eq:mstep-path} with the score form of \Cref{eq:mstep}, the parameter update of \citet{cuin2026learning}, under the $L_2$ and the PDE cost, with Holm corrections over the twelve settings of \Cref{tab:planners} (\Cref{app:ci}). With the PDE cost, the pathwise form is significantly ahead by $16$ to $25$ points on Reacher at $H{=}10$ and $H{=}15$, on Wall at $H{=}5$ and on Push-T at $H{=}5$ and $H{=}10$. After a Holm correction, the PDE cost is significantly ahead of the $L_2$ cost only under the pathwise form, on Reacher at $H{=}10$ and $H{=}15$; before it, also under the score form on Point maze at $H{=}5$ and $H{=}10$ and on Reacher at $H{=}15$. We believe that the differences between the two forms with the PDE cost are due to their estimators. \Cref{tab:abl-mstep} gives the remaining rows.

\begin{table}[!h]
  \caption{Success rate (\%) with the score form (\Cref{eq:mstep}) and pathwise form (\Cref{eq:mstep-path}) of the optimization step, each with the $L_2$ and PDE costs, on $100$ seeds per cell and $50$ on Cube; $^{\ast}$: as in \Cref{tab:planners,sec:exp:planning}; --: not run. $p$: exact McNemar test.}
  \label{tab:abl-mstep}
  \centering
  \scriptsize
  \setlength{\tabcolsep}{4pt}
  \begin{tabular}{llcccccccc}
    \toprule
    & & \multicolumn{2}{c}{Score form} & \multicolumn{2}{c}{Pathwise form} & \multicolumn{2}{c}{$p$, PDE vs.\ $L_2$} & \multicolumn{2}{c}{$p$, pathwise vs.\ score} \\
    \cmidrule(lr){3-4}\cmidrule(lr){5-6}\cmidrule(lr){7-8}\cmidrule(lr){9-10}
    Environment & $H$ & $L_2$ & PDE & $L_2$ & PDE & score & pathwise & $L_2$ & PDE \\
    \midrule
    Reacher & $5$ & 95 & 96 & 99 & 100$^{\ast}$ & 1.00 & 1.00 & 0.22 & 0.13 \\
    Reacher & $10$ & 72 & 84 & 78 & 100$^{\ast}$ & 0.050 & $<$0.001 & 0.36 & $<$0.001 \\
    Reacher & $15$ & 53 & 71 & 27 & 89$^{\ast}$ & 0.010 & $<$0.001 & $<$0.001 & 0.002 \\
    Point maze & $5$ & 79 & 88$^{\ast}$ & 87 & 81 & 0.049 & 0.31 & 0.057 & 0.065 \\
    Point maze & $10$ & 91 & 98$^{\ast}$ & 99 & 95 & 0.016 & 0.22 & 0.021 & 0.25 \\
    Point maze & $15$ & 95 & 96$^{\ast}$ & 91 & 91 & 1.00 & 1.00 & 0.34 & 0.23 \\
    Wall & $5$ & 75 & 75 & 89 & 92$^{\ast}$ & 1.00 & 0.45 & 0.001 & $<$0.001 \\
    Wall & $10$ & 84 & 84 & 89 & 81$^{\ast}$ & 1.00 & 0.12 & 0.27 & 0.68 \\
    Wall & $15$ & 89 & 79 & 91 & 84$^{\ast}$ & 0.087 & 0.092 & 0.82 & 0.42 \\
    Push-T & $5$ & 71 & 68 & 94 & 93$^{\ast}$ & 0.72 & 1.00 & $<$0.001 & $<$0.001 \\
    Push-T & $10$ & 18 & 15 & 35 & 32$^{\ast}$ & 0.61 & 0.69 & $<$0.001 & $<$0.001 \\
    Push-T & $15$ & 3 & 3 & 4 & 4$^{\ast}$ & 1.00 & 1.00 & 1.00 & 1.00 \\
    Cube & $5$ & 84 & -- & 86 & 88$^{\ast}$ & -- & 1.00 & 1.00 & -- \\
    \bottomrule
  \end{tabular}
\end{table}

\subsection{Move size.}\label{app:abl-sens} The reference width $\sigma_p$, the standard deviation of the reference $q(\cdot\mid\bm{\mu})$ of \Cref{eq:target_family}, sets how much ESS a move of the plan costs: at $\beta=0$, the ESS of the factors $q(\mathbf a\mid\bm{\mu}_k)/q(\mathbf a\mid\bm{\mu}_{k-1})$ approaches $N\exp\!\big(-\lVert\bm{\mu}_k-\bm{\mu}_{k-1}\rVert^2/\sigma_p^2\big)$ for large $N$ \citep[Sec.~4.2]{cuin2026efficientstochasticoptimisationsequential}. Holm corrections are over the seventeen changed settings of \Cref{tab:abl-sens} and the ten of \Cref{tab:abl-smc} in each cell (\Cref{app:ci}). Of the settings of \Cref{tab:abl-sens}, only $\sigma_p$ changes success significantly: $\sigma_p=0.5$ lowers it to $86\%$ on Point maze ($p=0.022$) and $\sigma_p=2$ to $55\%$ on Reacher ($p<0.001$), and only the latter survives a Holm correction. The population is resampled at $94\%$ to $97\%$ of the iterations with $\sigma_p=0.5$ and at $37\%$ to $45\%$ with $\sigma_p=2$, against $49\%$ to $51\%$ with $\sigma_p=1$ (\Cref{tab:stab-resample}). We have not determined why the narrower reference lowers success on Point maze and the wider one on Reacher.

\begin{table}[!h]
  \caption{Success (\%) with $95\%$ Wilson interval, mean iterations to success of successful seeds, and exact McNemar $p$-value against the default$^{\ast}$ (\Cref{tab:hyper}), at $H{=}10$ on $100$ seeds, one setting changed at a time; operators as in \Cref{tab:calib-operators}. Point maze rows use the pathwise optimization step.}
  \label{tab:abl-sens}
  \centering
  \scriptsize
  \setlength{\tabcolsep}{3.5pt}
  \begin{tabular}{llcccccc}
    \toprule
    & & \multicolumn{3}{c}{Point maze $H{=}10$} & \multicolumn{3}{c}{Reacher $H{=}10$} \\
    \cmidrule(lr){3-5}\cmidrule(lr){6-8}
    & Setting & Success (\%) & Iter. & $p$ & Success (\%) & Iter. & $p$ \\
    \midrule
    \multirow{5}{*}{Temperature $\alpha/\operatorname{std}C$} & 0.25 & 93 [86, 97] & 9.4 & 0.69 & 98 [93, 99] & 31.2 & 0.50 \\
     & 0.5 & 90 [83, 94] & 9.3 & 0.13 & 97 [92, 99] & 22.2 & 0.25 \\
     & 1$^{\ast}$ & 95 [89, 98] & 7.8 & -- & 100 [96, 100] & 25.2 & -- \\
     & 2 & 95 [89, 98] & 8.6 & 1.00 & 100 [96, 100] & 28.2 & 1.00 \\
     & 4 & 93 [86, 97] & 9.0 & 0.73 & 99 [95, 100] & 29.9 & 1.00 \\
    \midrule
    \multirow{3}{*}{Step size $\eta$} & 0.09 & 94 [88, 97] & 7.0 & 1.00 & 100 [96, 100] & 16.4 & 1.00 \\
     & 0.18$^{\ast}$ & 95 [89, 98] & 7.8 & -- & 100 [96, 100] & 25.2 & -- \\
     & 0.36 & 94 [88, 97] & 8.8 & 1.00 & 98 [93, 99] & 32.2 & 0.50 \\
    \midrule
    \multirow{3}{*}{Reference width $\sigma_p$} & 0.5 & 86 [78, 91] & 10.7 & 0.022 & 100 [96, 100] & 16.0 & 1.00 \\
     & 1$^{\ast}$ & 95 [89, 98] & 7.8 & -- & 100 [96, 100] & 25.2 & -- \\
     & 2 & 98 [93, 99] & 9.7 & 0.45 & 55 [45, 64] & 49.8 & $<$0.001 \\
    \midrule
    \multirow{3}{*}{Roadmap $k$} & $k/2$ & 90 [83, 94] & 7.8 & 0.23 & 97 [92, 99] & 22.2 & 0.25 \\
     & $k$$^{\ast}$ & 95 [89, 98] & 7.8 & -- & 100 [96, 100] & 25.2 & -- \\
     & $2k$ & 93 [86, 97] & 7.2 & 0.73 & 100 [96, 100] & 22.7 & 1.00 \\
    \midrule
    \multirow{3}{*}{Diffusion constant $c$} & $c/3$ & 92 [85, 96] & 9.5 & 0.45 & 99 [95, 100] & 24.5 & 1.00 \\
     & $c$$^{\ast}$ & 95 [89, 98] & 7.8 & -- & 100 [96, 100] & 25.2 & -- \\
     & $3c$ & 96 [90, 98] & 9.4 & 1.00 & 99 [95, 100] & 19.6 & 1.00 \\
    \midrule
    \multirow{3}{*}{PDE weight $\lambda_V$} & $\times0.5$ & 94 [88, 97] & 12.0 & 1.00 & 99 [95, 100] & 23.3 & 1.00 \\
     & $\times1$$^{\ast}$ & 95 [89, 98] & 7.8 & -- & 100 [96, 100] & 25.2 & -- \\
     & $\times2$ & 98 [93, 99] & 8.3 & 0.45 & 100 [96, 100] & 25.3 & 1.00 \\
    \midrule
    \multirow{4}{*}{Operator on the roadmap} & heat step$^{\ast}$ & 95 [89, 98] & 7.8 & -- & 100 [96, 100] & 25.2 & -- \\
     & number of edges & 94 [88, 97] & 8.2 & 1.00 & 100 [96, 100] & 24.6 & 1.00 \\
     & Green's function & 96 [90, 98] & 8.9 & 1.00 & 100 [96, 100] & 25.4 & 1.00 \\
     & shortest path & 93 [86, 97] & 8.0 & 0.73 & 100 [96, 100] & 23.5 & 1.00 \\
    \bottomrule
  \end{tabular}
\end{table}

\begin{table}[!h]
  \caption{Share of the iterations at which the population is resampled, in the two cells of \Cref{tab:abl-sens}. Each row below the first changes one setting of \Cref{tab:hyper}.}
  \label{tab:stab-resample}
  \centering
  \small
  \begin{tabular}{lcc}
    \toprule
    Setting & Reacher & Point maze \\
    \midrule
    settings marked $^{\ast}$ & $51\%$ & $49\%$ \\
    $\eta=0.09$               & $23\%$ & $32\%$ \\
    $\eta=0.36$               & $90\%$ & $86\%$ \\
    $\sigma_p=2$              & $37\%$ & $45\%$ \\
    $\sigma_p=0.5$            & $94\%$ & $97\%$ \\
    \bottomrule
  \end{tabular}
\end{table}

\subsection{Failure modes.}\label{app:abl-smc} \Cref{tab:abl-smc} varies the number $N$ of particles from $2$ to $32$. After a Holm correction, the ablations of \Cref{tab:abl-smc,tab:abl-sens} lower success significantly only on Reacher, to $39\%$, $85\%$ and $55\%$ with $N=2$, $N=4$ and $\sigma_p=2$ ($p<0.001$). On Point maze, $N=2$ and $\sigma_p=0.5$ lower it to $84\%$ and $86\%$, not significantly after the correction. We believe that the loss with $N=2$ is due to the rules of the planner, which estimate their quantities from the population: with two particles, the ESS never falls below $N/2$, so the population is never resampled and $\beta_k=1$ from the first iteration, and $\alpha$ and $\lambda_V$ are estimated from two particles. We have not isolated the cause of the losses with $N=4$ and with $\sigma_p=2$ on Reacher.

\begin{table}[!h]
  \caption{Population ablations of our planner at $H{=}10$ on $100$ seeds, one setting changed at a time from the default$^{\ast}$ (\Cref{tab:hyper}); columns as in \Cref{tab:abl-sens}. Point maze rows use the pathwise optimization step.}
  \label{tab:abl-smc}
  \centering
  \scriptsize
  \setlength{\tabcolsep}{3.5pt}
  \begin{tabular}{llcccccc}
    \toprule
    & & \multicolumn{3}{c}{Point maze $H{=}10$} & \multicolumn{3}{c}{Reacher $H{=}10$} \\
    \cmidrule(lr){3-5}\cmidrule(lr){6-8}
    & Setting & Success (\%) & Iter. & $p$ & Success (\%) & Iter. & $p$ \\
    \midrule
    \multirow{6}{*}{Particles $N$} & 2 & 84 [76, 90] & 18.8 & 0.007 & 39 [30, 49] & 51.9 & $<$0.001 \\
     & 4 & 91 [84, 95] & 11.2 & 0.29 & 85 [77, 91] & 47.4 & $<$0.001 \\
     & 8 & 93 [86, 97] & 11.2 & 0.77 & 96 [90, 98] & 28.0 & 0.13 \\
     & 12$^{\ast}$ & 95 [89, 98] & 7.8 & -- & 100 [96, 100] & 25.2 & -- \\
     & 16 & 97 [92, 99] & 9.0 & 0.50 & 100 [96, 100] & 19.1 & 1.00 \\
     & 32 & 94 [88, 97] & 7.7 & 1.00 & 100 [96, 100] & 12.7 & 1.00 \\
    \midrule
    \multirow{5}{*}{Resampling threshold $\tau$} & 0 (never) & 96 [90, 98] & 4.8 & 1.00 & 99 [95, 100] & 28.9 & 1.00 \\
     & 0.25 & 92 [85, 96] & 6.8 & 0.25 & 100 [96, 100] & 25.5 & 1.00 \\
     & 0.5$^{\ast}$ & 95 [89, 98] & 7.8 & -- & 100 [96, 100] & 25.2 & -- \\
     & 0.75 & 94 [88, 97] & 8.4 & 1.00 & 99 [95, 100] & 17.7 & 1.00 \\
     & 1 (always) & 93 [86, 97] & 7.8 & 0.75 & 99 [95, 100] & 20.0 & 1.00 \\
    \midrule
    \multirow{2}{*}{Weights} & SMC weights$^{\ast}$ & 95 [89, 98] & 7.8 & -- & 100 [96, 100] & 25.2 & -- \\
     & uniform & 98 [93, 99] & 9.3 & 0.38 & 100 [96, 100] & 25.8 & 1.00 \\
    \bottomrule
  \end{tabular}
\end{table}

\subsection{Remaining settings.} The temperature $\alpha$ ($0.25s$ to $4s$, with $s$ the standard deviation of $C$ over the initial population), the step size $\eta$ ($0.09$ to $0.36$), the resampling threshold $\tau$ ($0$, never, to $1$, always) and uniform weights, the number $\kappa$ of nearest neighbors of the roadmap ($\kappa/2$ to $2\kappa$), the diffusion constant $c$ ($c/3$ to $3c$; \Cref{app:impl,app:operator}), the PDE weight $\lambda_V$ (halved or doubled) and the operator on the roadmap (the number of edges, the Green's function or the shortest-path distance in place of the heat step of \Cref{eq:heat}) do not change success significantly in our settings over these ranges ($p\ge0.13$), and \Cref{tab:abl-sens,tab:abl-smc} summarize the numbers. Replacing the operator changes success by at most $2$ points in either cell. We believe that the absence of a gain from the settings of the cost is due to the ceiling of these cells, so these rows do not single out the heat step or the values of $\kappa$, $c$ and $\lambda_V$. ULA does not reject a step beyond the stability limit of its drift: for the curvature $\sigma_p^{-2}$ of the reference, $\eta/\sigma_p^2$ is at most $0.72$ in \Cref{tab:abl-sens}, below the limit $2$; we do not bound the curvature of $\beta_kC/\alpha$, but no run of our planner in \Cref{tab:planners,tab:abl-mstep,tab:abl-smc,tab:abl-sens} produced a non-finite cost.

\section{Confidence intervals and paired tests}
\label{app:ci}

We report $95\%$ Wilson score intervals \cite{wilson1942confidence} for success rates. Unlike the Wald interval of \citet[Fig.~5]{psenka2026parallelstochasticgradientbasedplanning}, they remain informative at $0\%$ and $100\%$. They are the error bars of \Cref{fig:kernels,fig:ablation}; the success curves are shown without bands. Since all arms of a comparison run on the same seeds, we compare two arms with the exact McNemar test, which uses only the seeds that exactly one of them solves. \Cref{tab:ci} gives the intervals for \Cref{tab:planners,tab:abl-1particle,tab:abl-cem} and Cube, and the $p$-value of our planner against the strongest baseline of \Cref{tab:planners} in each cell. All $p$-values are per comparison and unadjusted unless a Holm correction \cite{holm1979simple} within a family of comparisons is stated.

\paragraph{Planners.} Our planner is ahead of the strongest baseline in eight of the twelve cells of \Cref{tab:planners}, significantly at the $5\%$ level in four: on Reacher at every horizon and on Wall at $H{=}5$. No cell in which a baseline is ahead is significant ($p>0.28$). On Cube, with $50$ seeds, our planner is ahead of MPPI, GD and CEM without a significant difference ($p=1.00$, $0.73$ and $0.13$) and ahead of GRASP with $p<0.001$.

\paragraph{Holm correction.} Within each family of comparisons, the four significant cells of \Cref{tab:planners} survive the correction. So do the advantages of JALA over MALA on Point maze and over RWM on Reacher, Point maze, Wall and Push-T (\Cref{app:kernel-abl}), and of the PDE cost over the $L_2$ cost on Reacher at $H{=}10$ and $H{=}15$ (\Cref{sec:exp:ablation}). The advantage of JALA over MALA on Push-T does not, nor do those of the PDE cost on Point maze at $H{=}5$ and $H{=}10$ ($p=0.049$ and $p=0.016$ uncorrected). No other difference between the two costs is significant even without the correction.

\begin{table}[!h]
  \caption{$95\%$ Wilson intervals (\%) for \Cref{tab:planners} and Cube (top) and \Cref{tab:abl-1particle,tab:abl-cem} (bottom); $p$: exact McNemar test of ours against the strongest baseline.}
  \label{tab:ci}
  \centering
  \setlength{\tabcolsep}{5pt}
  \renewcommand{\arraystretch}{1.15}
  \resizebox{1\textwidth}{!}{%
  \begin{tabular}{lccccccccccccc}
    \toprule
    & \multicolumn{3}{c}{Reacher} & \multicolumn{3}{c}{Point maze} & \multicolumn{3}{c}{Wall} & \multicolumn{3}{c}{Push-T} & Cube \\
    \cmidrule(lr){2-4} \cmidrule(lr){5-7} \cmidrule(lr){8-10} \cmidrule(lr){11-13} \cmidrule(lr){14-14}
    Planner & 5 & 10 & 15 & 5 & 10 & 15 & 5 & 10 & 15 & 5 & 10 & 15 & 5 \\
    \midrule
    GD & 74--89 & 29--48 & 9--23 & 39--59 & 56--75 & 59--77 & 48--67 & 48--67 & 49--68 & 79--92 & 17--33 & 2--11 & 71--92 \\
    GRASP & 27--46 & 38--57 & 49--68 & 31--50 & 39--59 & 53--72 & 6--17 & 6--17 & 2--10 & 27--46 & 2--11 & 0--4 & 39--65 \\
    CEM & 59--77 & 33--52 & 32--51 & 49--68 & 66--82 & 71--87 & 21--39 & 32--51 & 46--65 & 33--52 & 3--12 & 1--8 & 65--87 \\
    MPPI & 78--91 & 69--85 & 57--75 & 73--88 & 89--98 & 92--99 & 69--85 & 79--92 & 78--91 & 74--89 & 19--36 & 1--7 & 74--93 \\
    \midrule
    \textbf{Ours} & 96--100 & 96--100 & 81--94 & 80--93 & 93--99 & 90--98 & 85--96 & 72--87 & 76--90 & 86--97 & 24--42 & 2--10 & 76--94 \\
    $p$, ours vs.\ strongest & $<$0.001 & $<$0.001 & $<$0.001 & 0.15 & 0.45 & 1.00 & $<$0.001 & 0.29 & 0.83 & 0.15 & 0.36 & 1.00 & 1.00 \\
    \midrule
    GD (1 particle) & 47--66 & 29--48 & 10--24 & 44--63 & 41--61 & 40--60 & 20--37 & 32--51 & 18--34 & 52--71 & 10--24 & 1--7 & \textendash \\
    GRASP (1 particle) & 19--36 & 25--43 & 24--42 & 35--54 & 39--59 & 52--71 & 14--30 & 5--16 & 2--11 & 28--47 & 2--10 & 0--4 & \textendash \\
    CEM ($30\times5$) & 52--71 & 29--48 & 20--37 & 41--61 & 57--75 & 59--77 & 25--43 & 31--50 & 39--59 & 36--55 & 2--10 & 1--7 & \textendash \\
    \bottomrule
  \end{tabular}}
\end{table}

\section{Per-seed comparisons}
\label{app:paired}
\label{app:matched}

\begin{figure}[!tbp]
  \centering
  \includegraphics[width=0.24\textwidth]{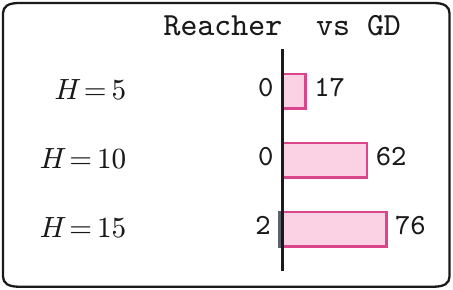} \hfill
  \includegraphics[width=0.24\textwidth]{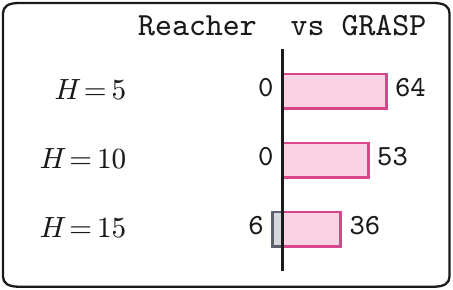} \hfill
  \includegraphics[width=0.24\textwidth]{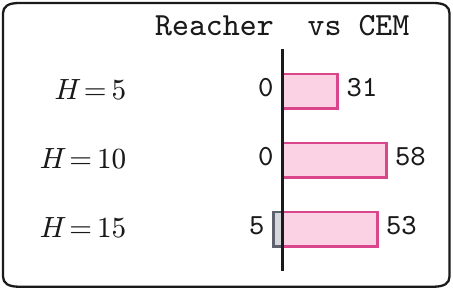} \hfill
  \includegraphics[width=0.24\textwidth]{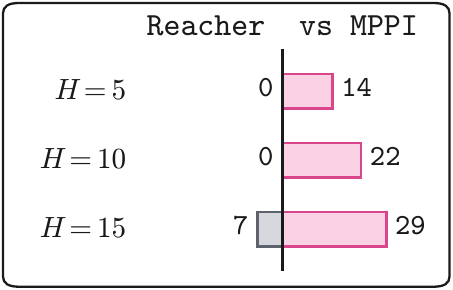} \\[4pt]
  \includegraphics[width=0.24\textwidth]{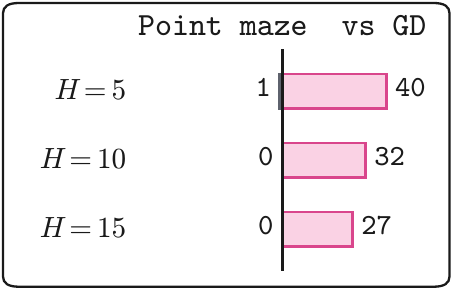} \hfill
  \includegraphics[width=0.24\textwidth]{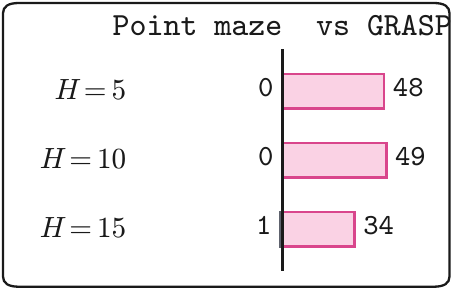} \hfill
  \includegraphics[width=0.24\textwidth]{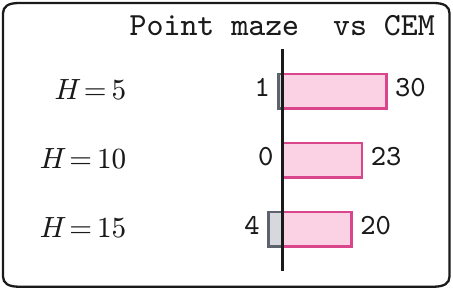} \hfill
  \includegraphics[width=0.24\textwidth]{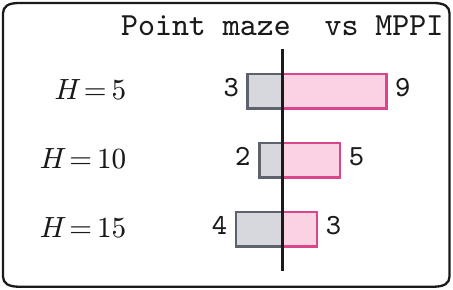} \\[4pt]
  \includegraphics[width=0.24\textwidth]{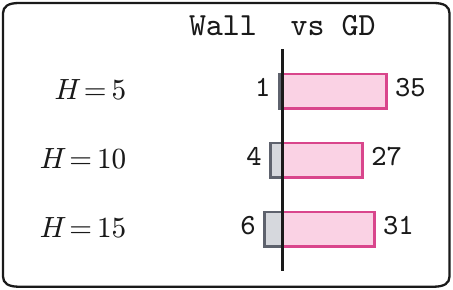} \hfill
  \includegraphics[width=0.24\textwidth]{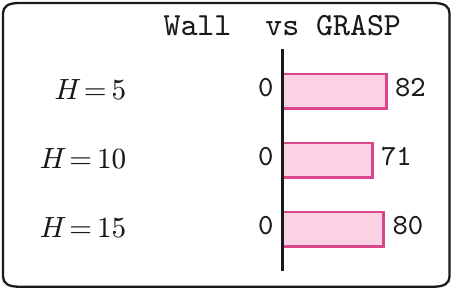} \hfill
  \includegraphics[width=0.24\textwidth]{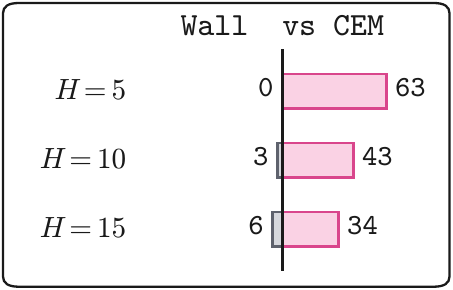} \hfill
  \includegraphics[width=0.24\textwidth]{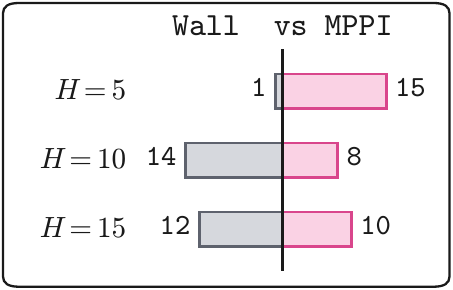} \\[4pt]
  \includegraphics[width=0.24\textwidth]{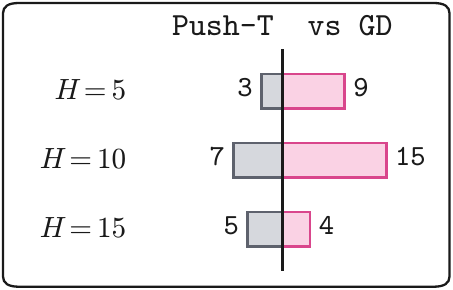} \hfill
  \includegraphics[width=0.24\textwidth]{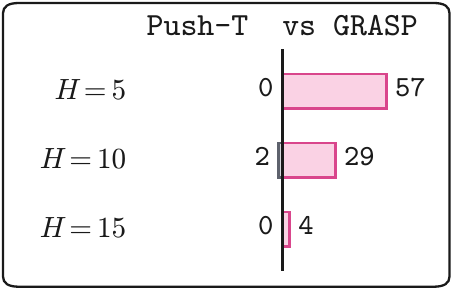} \hfill
  \includegraphics[width=0.24\textwidth]{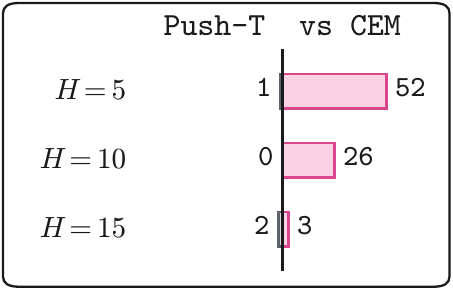} \hfill
  \includegraphics[width=0.24\textwidth]{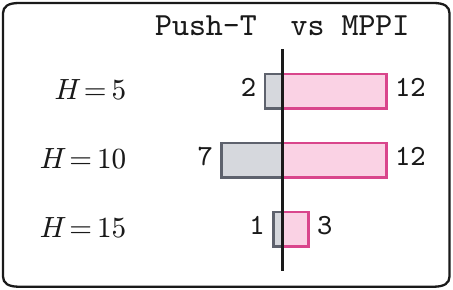}
  \caption{Per-seed paired outcomes against each baseline of \Cref{tab:planners}, one panel per environment and baseline, every planner with $12$ particles and $100$ seeds per setting. Bars to the right count seeds we solve and the baseline does not; bars to the left count the reverse. One-sided bars mean that one method solves a superset of the seeds of the other.}
  \label{fig:paired}
\end{figure}

\Cref{fig:paired} splits the paired outcomes of \Cref{fig:paired_summary} by environment.

\section{Success against iteration, remaining panels}
\label{app:curves}

\begin{figure}[!htbp]
  \centering
  \includegraphics[width=0.315\textwidth]{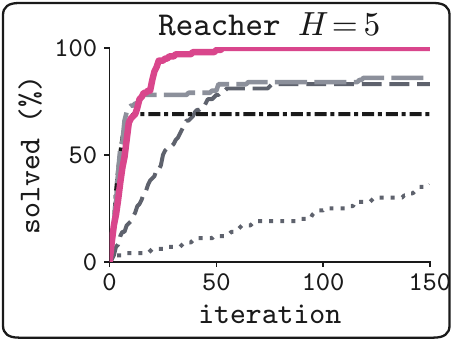} \hfill
  \includegraphics[width=0.315\textwidth]{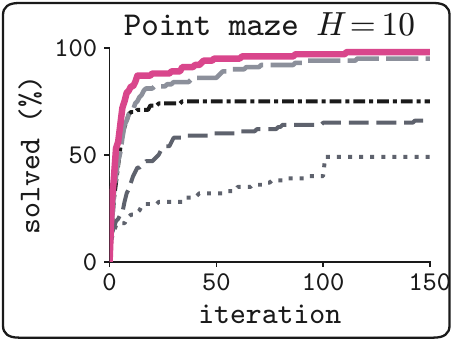} \hfill
  \includegraphics[width=0.315\textwidth]{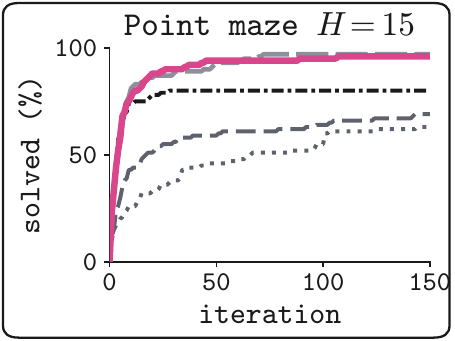} \\[4pt]
  \includegraphics[width=0.315\textwidth]{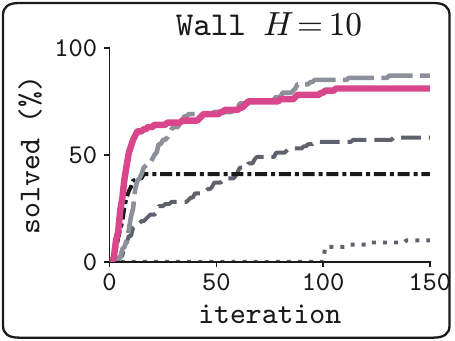} \hfill
  \includegraphics[width=0.315\textwidth]{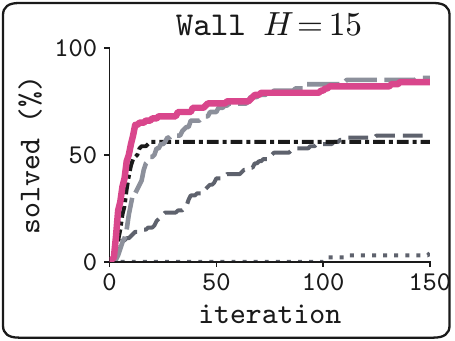} \hfill
  \includegraphics[width=0.315\textwidth]{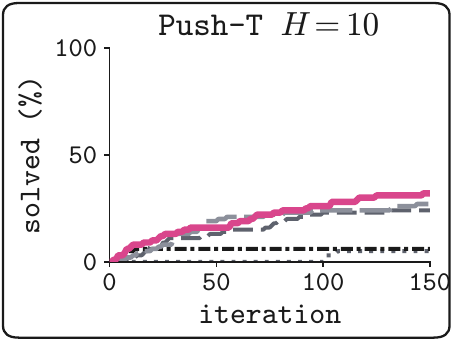} \\[4pt]
  \includegraphics[width=0.315\textwidth]{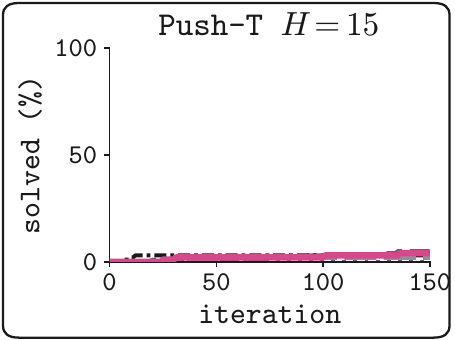} \\[2pt]
  \includegraphics[width=0.62\textwidth]{figs/fig_success_legend}
  \caption{Fraction of seeds solved at or before each iteration, for the settings not shown in \Cref{fig:success}.}
  \label{fig:curves_rest}
\end{figure}

\Cref{fig:curves_rest} shows the settings omitted from \Cref{fig:success}, so that the full grid is available without selecting on outcome.

\section{Rollouts against gradient descent with 1 particle and 12 candidates}
\label{app:frames}

\begin{figure}[!htbp]
  \centering
  \includegraphics[width=\linewidth]{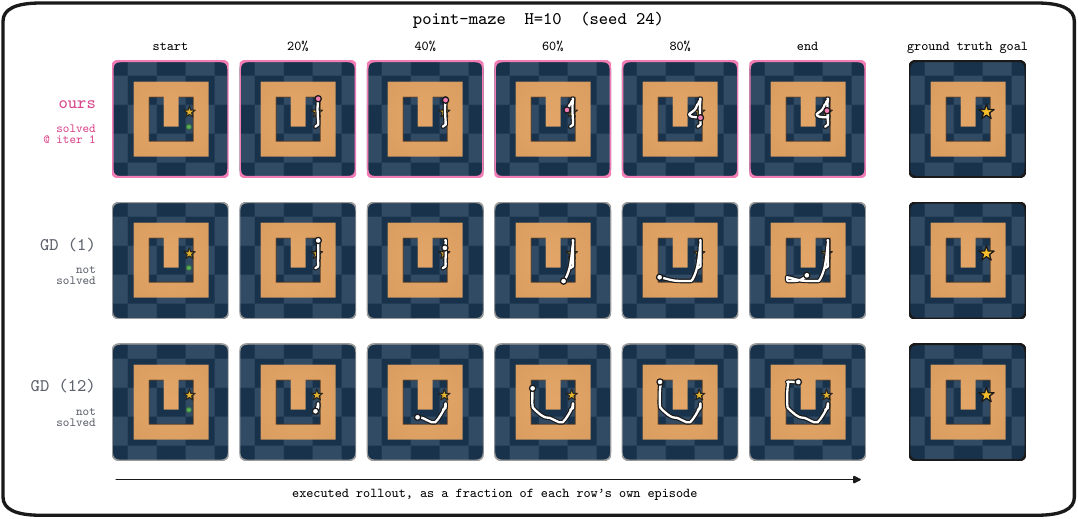}
  \caption{Point maze rollouts ($H{=}10$, seed 24) of our planner and of GD with a single particle and with $12$ candidates; the goal frame is at the right. Our planner, run here with the pathwise form of \Cref{eq:mstep-path} of the optimization step rather than the score form used on Point maze in \Cref{tab:planners}, solves in $1$ iteration; GD reaches the $150$-iteration cap in both settings.}
  \label{fig:frames_pm_app}
\end{figure}

\begin{figure}[!htbp]
  \centering
  \includegraphics[width=\linewidth]{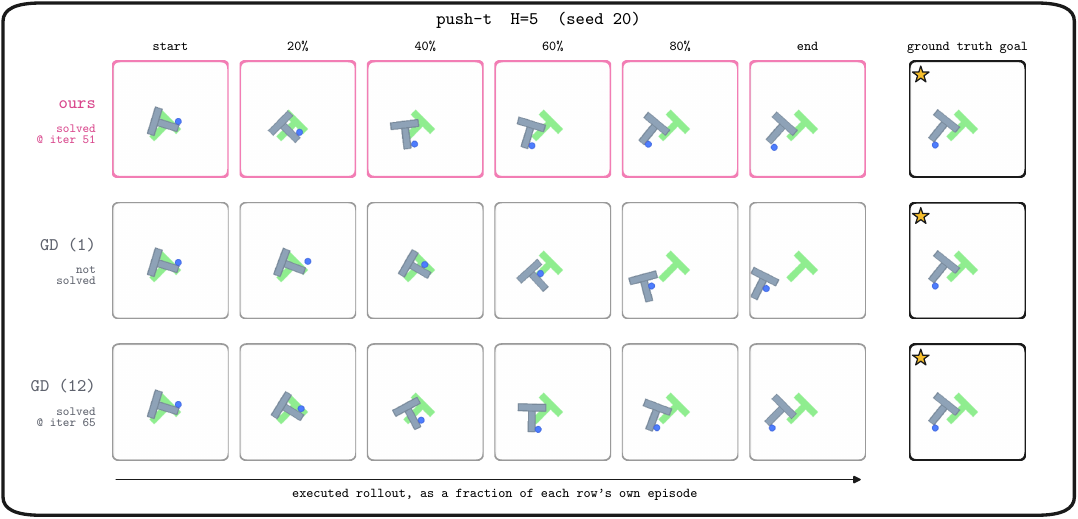}
  \caption{Push-T rollouts ($H{=}5$, seed 20) of our planner and of GD with a single particle and with $12$ candidates. Ours solves in $51$ iterations and GD with $12$ candidates in $65$; GD with a single particle reaches the $150$-iteration cap.}
  \label{fig:frames_pusht_app}
\end{figure}

\Cref{fig:frames_pm_app,fig:frames_pusht_app} compare rollouts on two further seeds with those of GD with a single particle (\Cref{app:abl-baselines}) and with the $12$ candidates of \Cref{tab:planners}, on Point maze at $H{=}10$ and Push-T at $H{=}5$.

\end{document}